\documentclass{article}

\usepackage{arxiv}
\usepackage[numbers]{natbib}

\usepackage[utf8]{inputenc} 
\usepackage[T1]{fontenc}    

\usepackage{amsmath,amsfonts,amsthm}
\newtheorem{theorem}{Theorem}[section]

\newtheorem{assumption}[theorem]{Assumption}

\usepackage{subcaption} 
\usepackage{caption}

\usepackage{tikz}
\usepackage{varwidth}
\usepackage{curves}
\usetikzlibrary{positioning,fit,calc,arrows,shapes,backgrounds,plotmarks}
\usepackage{pgfplots}
\pgfplotsset{compat=1.17}

\usepackage{booktabs}
\usepackage{enumitem}

\usepackage{multirow}
\newcommand{\norm}[1]{\left\lVert#1\right\rVert}
\usepackage[hyphens]{url}

\usepackage{verbatim}
\usepackage{adjustbox}
\usepackage{wrapfig}

\usepackage{algorithm}
\usepackage{algpseudocode}
\newcommand{\algorithmicbreak}{\textbf{break}}
\newcommand{\BREAK}{\State \algorithmicbreak}
\algnewcommand{\algorithmicand}{\textbf{ and }}
\algnewcommand{\algorithmicor}{\textbf{ or }}
\algnewcommand{\OR}{\algorithmicor}
\algnewcommand{\AND}{\algorithmicand}
\algnewcommand\algorithmicinput{\textbf{Input:}}
\algnewcommand\algorithmicoutput{\textbf{Output:}}
\algnewcommand\Input{\algorithmicinput}%
\algnewcommand\Output{\algorithmicoutput}%

\usepackage{cleveref}       

\title{Classification with Trust: A Supervised Approach based on Sequential Ellipsoidal Partitioning}

\author{
  Ranjani Niranjan \\
  International Institute of Information Technology Bangalore \\
  Bangalore, India\\
  \texttt{ranjani.niranjan@iiitb.ac.in} \\
   \And
  Sachit Rao \\
  International Institute of Information Technology Bangalore \\
  Bangalore, India\\
  \texttt{sachit@iiitb.ac.in}
}

\renewcommand{\shorttitle}{Classification by Sequential Ellipsoidal Partitioning}

\begin{document}
\maketitle

\begin{abstract}
Standard metrics of performance of classifiers, such as accuracy and sensitivity, do not reveal the trust or confidence in the predicted labels of data. While other metrics such as the computed probability of a label or the signed distance from a hyperplane can act as a trust measure, these are subjected to heuristic thresholds. This paper presents a convex optimization-based supervised classifier that sequentially partitions a dataset into several ellipsoids, where each ellipsoid contains nearly all points of the same label. By stating classification rules based on this partitioning, Bayes' formula is then applied to calculate a trust score to a label assigned to a test datapoint determined from these rules. The proposed Sequential Ellipsoidal Partitioning Classifier (SEP-C) exposes dataset irregularities, such as degree of overlap, without requiring a separate exploratory data analysis. The rules of classification, which are free of hyperparameters, are also not affected by class-imbalance, the underlying data distribution, or number of features. SEP-C does not require the use of non-linear kernels when the dataset is not linearly separable. The performance, and comparison with other methods, of SEP-C is demonstrated on the XOR-problem, circle dataset, and other open-source datasets. 
\end{abstract}

\keywords{Supervised Classification, Prediction with Trust, Ellipsoidal approximation, Convex Optimization}

\section{Introduction}\label{sec:introduction}

With the availability of labelled data, supervised-learning based classifiers have found applications in many domains. The performance of the classifier is dependent on the characteristics of the dataset itself, such as class imbalance, overlap, and outliers; such characteristics are usually identified by the additional process of Exploratory Data Analysis (EDA). In this paper, we present a classifier that simultaneously reveals the nature of the dataset, without the need of an EDA, and determines the trust, or ``confidence measure'' as denoted in the literature, in the prediction of a label by the classifier. Generating a trust score enables a user to override/accept the classifier's output, as required in Active Learning applications.

In the proposed classifier, which we term as the Sequential Ellipsoidal Partitioning Classifier (SEP-C), we first determine a hyperplane, akin to that in Support Vector Machines (SVMs), to separate datapoints from the training dataset of different labels; second, iteratively determine \textit{non-overlapping} Minimum Volume Ellipsoids (MVEs), \cite{boyd2004convex,Sun2004,Kong2007}, that cover these points, using the Reduced Convex Hull (RCH) Algorithm, \cite{bennett2000duality}; and third, remove the covered points from the dataset. This process is carried out until no points remain uncovered in the dataset or the independence-dimension inequality is violated, \cite{boyd2018linalg}. It is highlighted that in each iteration, the RCHs themselves are found iteratively and is not dependent on a single user-defined choice, as discussed in \cite{bennett2000duality}. The removal of points in each iteration renders this approach as a sequential one and in each iteration, a different hyperplane is found. Once the training dataset is partitioned into several MVEs, the trust in the predicted label of a test datapoint is calculated based on classification rules and Bayes' formula. These rules involve determining in which ellipsoid, or ellipsoids as the case may be, a test points lies; if a test point does not lie in any of the ellipsoids, the ellipsoid closest to it is expanded to contain it and a rule is applied. Consider a binary classification case and let a test point be contained in a single ellipsoid, say $\mathcal{E}_T$. Applying Bayes' formula, the resulting trust score, which is a conditional probability, is conditioned on the number of training set points of both labels that are contained in $\mathcal{E}_T$. Thus, if $\mathcal{E}_T$ contains majority of points of one label, then the trust in the prediction of that label is high. On the other hand, if the dataset has significant overlap and a test point lies in ellipsoid(s) that contain nearly equal number of points of both labels, then, clearly the trust in predicting either label is close to 50\%; the trust score determined this way can aid the user in abstaining from assigning a label to that point. While SEP-C emphasises the formation of non-overlapping MVEs, if the dataset is heavily overlapped, then, the partitioning algorithm may terminate in the first iteration itself, thus making it similar to the conventional SVM. In such cases, SEP-C allows for a user-defined number of points of one label to be contained in the MVE of points of another label; this number is the lone design parameter of SEP-C. With this approach, as we show, points that are ``far'' from the overlapping region can be assigned a label with a high trust. 

Conformal prediction has been suggested as a method to measure the confidence in the output of a classifier, \cite{shafer2008tutorial,balasubramanian2014conformal}. In this approach, prediction regions that contain the true label of a test data point with probability $(1-\epsilon)$ are defined. The generation of such regions is based on calculating a nonconformity measure, which essentially is a measure of how ``different'' the test data point is when compared with the observed data points. For a binary SVM, the nonconformity measure is calculated using the widest interval that separates the 2 classes with minimum errors. Next, depending on where the test data point lies with respect to this interval, one of the nonconformity measure values is assigned to this data point. A probability value, $p_y$, is then calculated using nonconformity measure values calculated for all the observed data points. If $p_y>\epsilon$, then, the confidence of the predicted label is assigned as $(1-\epsilon)$. This technique can result in regions being empty or containing multiple labels. In both cases, the calculated confidence measure does not yield the desired trust. In \cite{ribeiro2016should}, an \textit{explainer} model is demonstrated that can be used to explain the results of any classifier; this method uses a distance-metric and a function that represents the classifier.

We now review results in the literature which share similarities with one or more components of SEP-C. In the context of overlapping datasets with/without outliers, hyperplane-based methods are designed under the assumption of linear separability or no-overlap; this approach is prone to classification errors. Although the classical SVM can handle nonlinear structures in the dataset with the use of kernel trick, it is a single hyperplane that is used for classification; our approach, where multiple linear hyperplanes are defined, can be used for datasets that are not linearly separable. A variant of SVMs, denoted as the Ellipsoid SVM, is presented in \cite{Yao2008}, where MVEs that cover points belonging to the same label are found that are also separated by some margin; the ESVM finds a single ellipsoid per label and requires the kernel trick for certain datasets, while SEP-C finds several MVEs and does not require the use of any kernel. A strategy adopted to minimise the effects of overlap is to remove overlapping features or by merging them, \cite{saez2019addressing}. In \cite{elizondo2012linear}, it is suggested that more than one hyperplane is required to solve the classification problem when data is not linearly separable, though no mathematical proofs have been presented to support this. Outliers can be removed from datasets using methods suggested in  \cite{wang2019progress,erfani2016high,staerman2019functional}. Distance and density-based methods of outlier detection indicate that a single hyperplane classifier is unable to detect outliers efficiently. Trust in the outputs of SVMs has also been considered in the literature. Trust is computed by converting the SVM output to a posterior probability. In \cite{Madevska-Bogdanova:2004:PSO:2798096.2798147}, a distance metric, calculated as the output of a sigmoid function, is used to determine the posterior probability; a variant of this method is described in \cite{luo2005active}. A modified SVM classifier that can reject its output if the predicted output satisfies some conditions is presented in \cite{fumera2002support}.  For Active Learning, \cite{li2006confidence}, a histogram-based binning technique is suggested that converts the SVM output into a posterior probability; SVM margins and a confidence factor for this application are proposed in \cite{mitra2004probabilistic}. It is again highlighted that all these approaches use a single hyperplane.

Sequential ellipsoidal approaches have been presented, mainly for clustering or outlier detection, \cite{hodge2004survey}. A ``difference-of-squares'' minimisation method is proposed in \cite{Calafiore2002} that performs ellipsoidal fitting to points in $n$-dimensional space. An algorithm, denoted as the Bubble Clustering algorithm is presented in \cite{kraning2009bubble} that aims to cover points as a union of $k$ MVEs; a similar problem is addressed in \cite{MR2013}, where a dataset is partitioned into $E$ ellipsoids. It is highlighted that in both these approaches, $k$ and $E$ are user-specified parameters. In SEP-C, there is no need to specify the number of ellipsoids; they are found as a consequence of the algorithm. The only user-defined input is the number of misclassifications that are permitted while separating points of different labels - the first step of SEP-C. In an application of cell segmentation, \cite{PANAGIOTAKIS2020103810}, an algorithm that approximates any 2-dimensional shape using ellipses is presented. This algorithm does not require the number of ellipses to be specified \textit{a priori}; they are determined based on the degree of overlap of multiple ellipses - those that have a large overlap and do not contribute to additional information are removed. Ellipsoidal partitioning is analogous to approximating the distribution of any dataset with the use of finite Gaussian mixture models, \cite{McLachlan2019,Ghojogh2019}. The notable use of Gaussians in classification is in the Bayesian classifier, \cite{John2013}, where each attribute is modeled as a mixture of Gaussians. 

As SEP-C can be used as a basis to abstain from making predictions, this feature is markedly different than tree-based induction rule techniques, where, if a test point belongs to a leaf that contains equal number of points of both labels, this test point is assigned the majority class label. The use of such inductive algorithms is adopted in \cite{martens2009} to extract rules in SVMs, where, labels of points misclassified by a trained SVM is changed and next, new points are generated in the vicinity of the support vectors; see \cite{martens2009} of other rule-generation based methods for SVMs, such as \cite{4161896,Fung2008}, and their challenges. It is noted that the mere stating of rules does not indicate the trust in prediction of a test datapoint. The motivation with these approaches is to reduce the ``black box'' characteristics of SVM-type classifiers; we believe that classification from ellipsoidal partitioning can answer the question of ``why this label?'', as predicted by SEP-C.

The main contributions of SEP-C are that it \textit{i}. uses convex methods at every step, and hence, the problem of local minima is avoided; \textit{ii}. is free of hyperparameters, say as used in the SVM cost function; \textit{iii}. implicitly contains multiple hyperplanes, thus allowing for non-linearly separable datasets to be handled; \textit{iv}. presents a landscape of the dataset, for instance, by determining the degree of overlap of points of more than one label; and \textit{v}. calculates trust in prediction, that is based on finding points of the same label in the vicinity of the test point. SEP-C can also handle multi-class datasets by adopting the one-vs-all approach. We also believe that SEP-C can be used for data generation and the identification of small disjuncts, but we defer this to future work. As we show in the Results section, ellipsoidal partitioning is quite similar to the regions formed by the rules that lead to leaf nodes, from the root, in Decision Trees (DTs); however, the regions in DTs can be unbounded, while the ellipsoids are not.

\section{Preliminaries}\label{Sec:Prelim}

In this section, the problem statement, ellipsoidal representations, and the key convex optimization problems (CPs) and their solutions that form the main components of SEP-C are briefly described. The conditions that lead to these CPs being feasible are also highlighted.

\subsection{Problem Statement}\label{Sec:PobStat}

\begin{figure}[htb!]
    \centering
    \hspace{-9mm}
    \begin{subfigure}[b]{0.48\columnwidth}
    \centering
        \includegraphics[scale = 0.26]{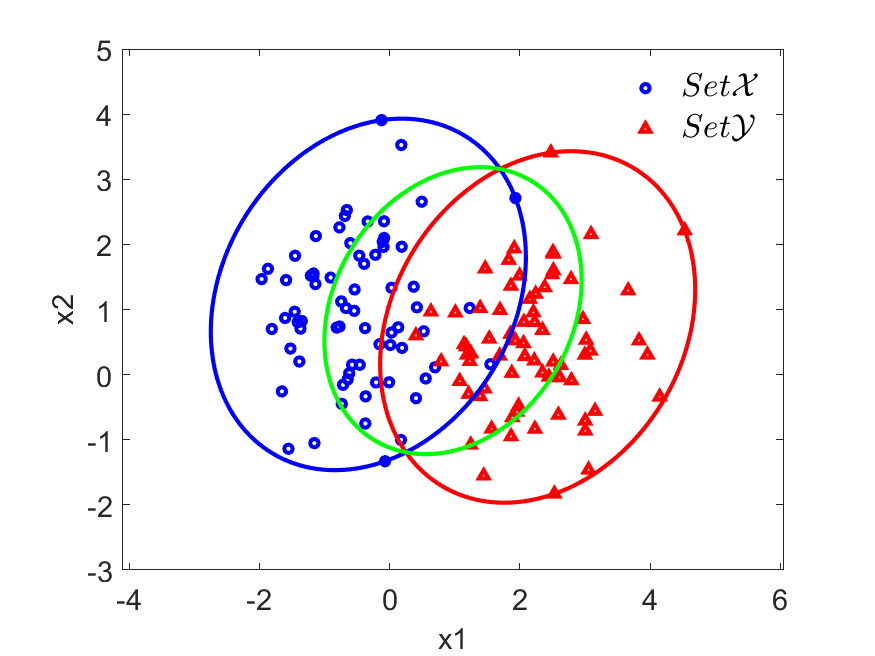}
        \caption{\label{Fig:Init_Ellps}}
    \end{subfigure} 
    \begin{subfigure}[b]{0.48\columnwidth}
    \centering
        \includegraphics[scale = 0.26]{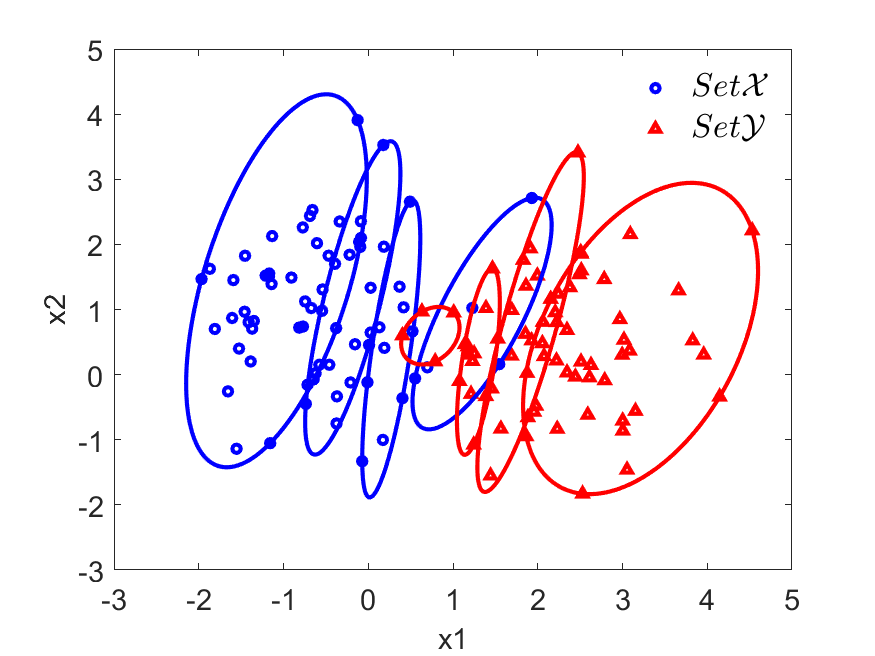}
        \caption{\label{Fig:EPart_Ov}}
    \end{subfigure}
    \caption{\ref{Fig:Init_Ellps}: 2-D Synthetic Overlapping Dataset; \ref{Fig:EPart_Ov}: Ellipsoidal partitioning using SEP-C}
    \label{fig:EllPart_Ov}
\end{figure}

The classifier proposed in this paper focuses on developing rules of classification for a dataset with points denoted by the sets $\mathcal{X}=\{x_i\}, \ x_i\in\Re^n, \ i=1,\cdots,N$ and $\mathcal{Y}=\{y_j\}, \ y_j\in\Re^n, \ j=1,\cdots,M$, where $N\geq M>2$, the points in set $\mathcal{X}$ have label $L_{+1}$, and those in $\mathcal{Y}$ have label $L_{-1}$. Note that a multi-label classification problem can be expressed as a binary classification problem using the one-vs-all approach. Similar to most classifiers, a part of the dataset is used to ``train'' the classifier and the remaining for testing and validation. The sets $\mathcal{X}$ and $\mathcal{Y}$ can overlap, such as the 2-dimensional (2-D) synthetic dataset shown in Fig.~\ref{Fig:Init_Ellps}, where the points marked in blue, say belong to $\mathcal{X}$, and the ones in red belong to $\mathcal{Y}$. 

\subsection{Ellipsoidal Approximation of a Dataset}\label{Sec:EllDataSet1}

For the sets $\mathcal{X}$ and $\mathcal{Y}$ shown in Fig.~\ref{Fig:Init_Ellps}, MVEs, denoted by $\mathcal{E}_X$ and $\mathcal{E}_Y$, respectively, can be found such that each ellipsoid contains points of the respective set either in its interior or its boundary; these ellipsoids are also known as the L\"{o}wner-John ellipsoids. An ellipsoid can be expressed in the form $\mathcal{E} = \{z \ | \ \norm{\mathbf{A}z+\mathbf{b}}\leq 1\}, \ z\in\Re^n$, where $\mathbf{A}\in\Re^{n\times n}$ is a positive-definite matrix and $\mathbf{b}\in\Re^n$. An alternative representation of this ellipsoid is given by the quadratic form resulting from expanding the 2-norm used in $\mathcal{E}$.

Now, given $N$ such points $z_i$, the MVE that covers these points, which is unique, is found as the solution to the convex optimisation problem (CP)
\begin{align}\label{Eq:MVE}
    &\min && \log \text{det}\left(\mathbf{A}^{-1}\right)  \\
    &\text{subject to } && \norm{\mathbf{A}z_i+\mathbf{b}}\leq 1, \ i=1,\cdots,N,\nonumber
\end{align}
with $\mathbf{A}$ and $\mathbf{b}$ as the variables, \cite{boyd2004convex}. It is highlighted that for the CP \eqref{Eq:MVE} to be feasible, the independence-dimension (I-D) inequality has to be satisfied, that is, $N>n$, \cite{boyd2018linalg}. The MVEs $\mathcal{E}_X$ and $\mathcal{E}_Y$, found by solving \eqref{Eq:MVE} are shown in Fig.~\ref{Fig:Init_Ellps} by the blue and red lines, respectively.

It can be seen that $\mathcal{E}_X$ and $\mathcal{E}_Y$ intersect since the points of both labels overlap with each other. The region of intersection can be approximated by an outer ellipsoid, denoted by $\mathcal{E}_{\text{Int}}=\{v \ | \ \norm{\mathbf{A}_{\text{Int}}v+\mathbf{b}_\text{Int}}\leq 1\}$ by solving the CP
\begin{subequations}\label{Eq:MVEIntEj}
    \begin{align}
        &\min && \log \text{det}\left(\mathbf{A}_\text{Int}^{-1}\right) \label{MVEIntEjCost}  \\
        &\text{subject to } && \begin{bmatrix}\mathbf{A}_\text{Int} & \mathbf{b}_\text{Int} & 0 \\ \mathbf{b}_\text{Int}^T & -1 & \mathbf{b}_\text{Int}^T \\ 0 & \mathbf{b}_\text{Int} & -\mathbf{A}_\text{Int}\end{bmatrix} \leq \sum\limits_{j=X,Y}{\tau_j} \begin{bmatrix}\mathbf{A}_j & \mathbf{b}_j & 0 \\ \mathbf{b}_j^T & c_j & 0 \\ 0 & 0 & 0\end{bmatrix}\label{MVEIntEjLMI} \\
        & && \tau_1\geq 0, \cdots, \tau_j\geq 0, \nonumber
    \end{align}
\end{subequations}
where, $\mathbf{A}_j,\mathbf{b}_j$, and $c_j$ are given by the \textit{quadratic representations} of the the ellipsoids $\mathcal{E}_j, \ j=X,Y$. In \eqref{Eq:MVEIntEj}, the constraints \eqref{MVEIntEjLMI} are Linear Matrix Inequalities (LMIs). Intersection of ellipsoids can also be determined by the algorithm presented in \cite[Sec.~15]{pope2008algorithms}, where a different representation of an ellipsoid is used. The result in \cite{pope2008algorithms} is connected to the Separating Hyperplane Theorem, which states that for two disjoint convex sets, a hyperplane that separates these sets can be found. Thus, if the ellipsoids, which are convex, indeed satisfy the property $\mathcal{E}_X \cap \mathcal{E}_Y = \emptyset$, then, a hyperplane that \textit{strictly} separates these ellipsoids exists; this hyperplane now acts as a classifier. 

The properties of $\mathcal{E}_{\text{Int}}$, such as its volume as well as the number of points contained within it, indicate the degree of overlap between the two sets $\mathcal{X}$ and $\mathcal{Y}$. If this ellipsoid is dense as well as ``large'', then, it can be expected that many points that have different labels are close to each other. The ellipsoid $\mathcal{E}_{\text{Int}}$ is marked in green in Fig.~\ref{Fig:Init_Ellps}. As can be expected, it is hard to classify the points that lie in the intersection region accurately. The CP \eqref{Eq:MVEIntEj} is solved using the CVX solver, \cite{cvx}, and the resulting solution is a ball for $\mathcal{E}_{\text{Int}}$, which is clearly an over-approximation of the overlapping region. A better approximation can be found by finding the MVE that covers the points that are contained both in $\mathcal{E}_X$ and $\mathcal{E}_Y$. It should be noted that the points in the overlapping region should again satisfy the I-D inequality, while the CP \eqref{Eq:MVEIntEj} can be applied even if the ellipsoids do intersect, but the region of overlap is empty.

SEP-C is in fact developed to partition such overlapping datasets into smaller, and possibly non-intersecting, ellipsoids, each of which contain points of the same label. The result of such partitioning using SEP-C is shown in Fig.~\ref{Fig:EPart_Ov}.

\subsection{Reduced Convex Hull Algorithm}\label{Sec:RCHull}

The RCH algorithm, \cite{bennett2000duality}, that forms a primary component of SEP-C is briefly described. The main motivation to apply the RCH algorithm is its ability to minimise the influence of outliers and overlap in classification. By defining the matrices $\mathbf{X}\in\Re^{N\times n}$ and $\mathbf{Y}\in\Re^{M\times n}$, where, each row of these matrices contains the datapoints $x_i$ and $y_j$, respectively, the RCHs of the sets $\mathcal{X}$ and $\mathcal{Y}$ are the set of all convex combinations $\mathbf{c}=\mathbf{X}^T\mathbf{u}$ and $\mathbf{d}=\mathbf{Y}^T\mathbf{v}$, respectively, where, $\mathbf{u}=[u_i]\in\Re^n, \ \mathbf{v}=[v_i]\in\Re^n$; $\sum{u_i}=1,\ 0\leq u_i \leq D$, $\sum{v_i}=1,\ 0\leq v_i \leq D$; and the scalar $D<1$. If the sets are balanced, that is, $N=M$, the parameter $D$ is chosen as $D=(1/K)$, where, $K\leq N$; in \cite{bennett2000duality}, $D$ is a design parameter. With the assumption that the RCH is non-empty for some $D<1$, the RCH algorithm finds the closest points in each RC-Hull by solving the CP
\begin{align}\label{Eq:RCH}
    &\min _{\mathbf{u},\mathbf{v}} && \frac{1}{2} \norm{\mathbf{X}^T\mathbf{u} -\mathbf{Y}^T\mathbf{v}}^2 \\
    &\text{subject to } && \mathbf{e}^T\mathbf{u} = 1,\ \mathbf{e}^T\mathbf{v} = 1, \ 0 \leq \mathbf{u,v} \leq D\mathbf{e}.\nonumber
\end{align}
The vector $\mathbf{e}^T=[1\ 1 \ \cdots]$. Now, if the solution to this CP exists for some $D<1$, the RC-Hulls do not intersect and thus, the line normal to the line connecting the closest points is the separating hyperplane.

\begin{figure}[htbp!]
    \centering
    \hspace{-8mm}
    \begin{subfigure}[b]{0.48\columnwidth}
    \centering
        \includegraphics[scale = 0.32]{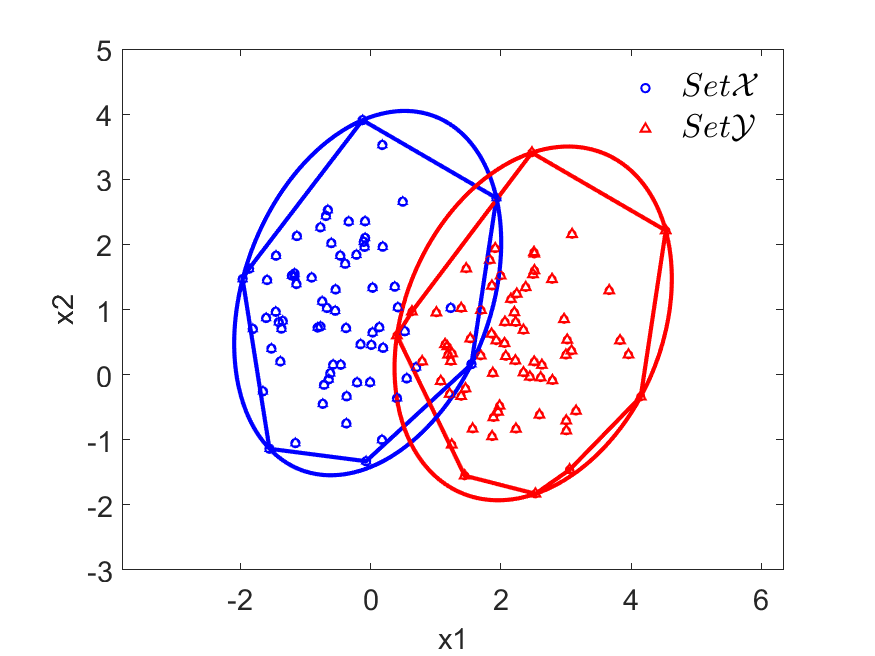}
        \caption{\label{fig:Init_CH_Ellp}}
    \end{subfigure} 
    \begin{subfigure}[b]{0.48\columnwidth}
    \centering
        \includegraphics[scale = 0.32]{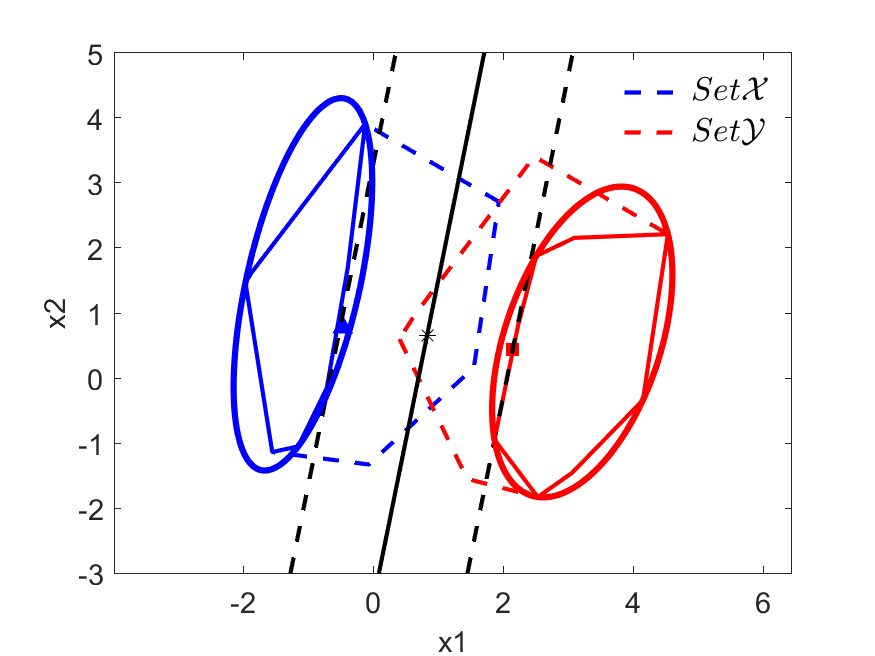}
        \caption{\label{fig:RCH_it1}}
    \end{subfigure}  
    
    \caption{\ref{fig:Init_CH_Ellp}: Intersecting CHs, and MVEs, of the two sets; \ref{fig:RCH_it1}: RCHs, and MVEs, that are non-intersecting (solid lines)}
    \label{fig:RCH_HP}
\end{figure}

Consider the 2-D synthetic dataset with two labels as shown in Fig.~\ref{fig:Init_CH_Ellp}; the convex hull and the MVE of each set are also shown. Solving the CP \eqref{Eq:RCH} results in the RCHs for the two sets, as shown in Fig.~\ref{fig:RCH_it1}. As can be seen, the original CHs and MVEs of both sets of points intersect, while the RCHs, found for some value of $D$, do not. 

As discussed in \cite{bennett2000duality}, the RCH algorithm is the dual of widening the margin in the classical SVM; this widening is performed to permit classification errors. Another similarity of the RCH approach is with the $\nu$-SVM classifier, \cite{nuSVMTutorial}, where the user-defined parameter $\nu\in[0,1]$ influences the cost function that is minimised and also has an upper-bound that is related to number of datapoints and SVs that are misclassified.

In SEP-C, the RCH algorithm is implemented differently. In place of the user-defined parameter $D=(1/K)$, beginning with $K=\min(N,M)$, the CP \eqref{Eq:RCH} is solved iteratively, where $K$ is reduced in each iteration, until such time that the RCHs of both sets do not intersect or the RCH of points of one label contain a few points of the other; the choice of number of such points is clarified in the section on description of SEP-C. We demonstrate that this approach leads to a trust score calculation with better interpretability and leads to some degree of partitioning in heavily overlapping datasets. In addition, experiments on synthetic datasets (not presented in this paper) indicates that this approach allows for imbalanced datasets to be handled. Further, to determine that the RCHs indeed do not intersect, the check on intersection is performed on the MVEs that cover them. The RCHs intersect if their respective MVEs intersect. This approach is selected to minimise computational cost, especially for high-dimensional datasets. 

\section{Main Results: SEP-C}

As mentioned in the Introduction, SEP-C is based on partitioning the training dataset into ellipsoidal regions, such that, \textit{as much as possible}, each ellipsoid contains points with the same label. With the identification of such regions, the classification rule for a new test point is simply to identify which ellipsoid it lies in and predict the label of the test point based the number of training points of either label that are contained in that ellipsoid. With this basic idea, first, the method followed to find such ellipsoidal regions is described. Next, the classification rule accompanied by a trust or ``odds'' measure is presented. 

\subsection{Ellipsoidal Partitioning} \label{Sec:EllPartition}

The ellipsoidal partitioning component of SEP-C is described in Algorithm~\ref{alg:EllipPart}; the notation $|\mathcal{X}|$ denotes the number of points in the set $\mathcal{X}$ and $\mathcal{X}\setminus \mathcal{Y}$ denotes the difference of sets $\mathcal{X},\mathcal{Y}$. This algorithm invokes the RCH algorithm, which is presented in Algorithm~\ref{alg:RCH}. Alg.~\ref{alg:EllipPart} consists of two user-defined parameters: the integers $n>0$, which denotes the dimension of the feature space, and $n_\text{Imp}\geq0$, which denotes the number of permitted misclassifications. The roles of these parameters are clarified in the following sections.

\begin{algorithm}
    \caption{Ellipsoidal Partitioning}\label{alg:EllipPart}
    \begin{algorithmic}[1]
    
        \State $i \gets 1$
        \State $\mathcal{E}_X,\mathcal{E}_Y\gets$ MVEs of $\mathcal{X}$ and $\mathcal{Y}$
        \State $\mathcal{X}^+,\mathcal{X}^-,\mathcal{Y}^+,\mathcal{Y}^-\gets\emptyset$
        \State $\mathcal{E}^+,\mathcal{E}^-\gets\emptyset$
        \State $n_{\text{Imp}}\geq0$ \Comment{Number of points of one label allowed in the set of another}
        \State $n>0$ \Comment{Dimension of the feature space}
        \While{$|\mathcal{X}| > n$ \AND $|\mathcal{Y}| > n $} \Comment{Ensure I-D condition is satisfied}
            
            \State $\left(\mathcal{X}^+_i,\mathcal{Y}^-_i,\mathcal{E}_{X^+},\mathcal{E}_{Y^-}\right)\gets \text{RCH}\left(\mathcal{X},\mathcal{Y}\right)$ \Comment{RCH Algorithm}
            
            \If{$\mathcal{X}^+_i=\emptyset$ \OR $\mathcal{Y}^-_i=\emptyset$ } 
                \State Datasets $\mathcal{X}$ and $\mathcal{Y}$ cannot be separated further
                \BREAK
            \Else
                \While{$|\mathcal{Y}\in \mathcal{E}_{X^+} | \geq n_{\text{Imp}}$}
                    \State $\mathcal{X}^+_i\gets\text{RCH}\left(\mathcal{X}^+_i,\mathcal{Y}\right)$ \Comment{Subset of $\mathcal{X}$ containing no greater than $n_{\text{Imp}}$ points of $\mathcal{Y}$}
                \EndWhile
                \While{$|\mathcal{X}\in \mathcal{E}_{Y^-} | \geq n_{\text{Imp}}$}
                    \State $\mathcal{Y}^-_i\gets\text{RCH}\left(\mathcal{Y}^-_i,\mathcal{X}\right)$ \Comment{Subset of $\mathcal{Y}$ containing no greater than $n_{\text{Imp}}$ points of $\mathcal{X}$}
                \EndWhile
                
                \State $\mathcal{X}^+\gets\{\mathcal{X}^+_i\}, \ \mathcal{Y}^-\gets\{\mathcal{Y}^-_i\}$
                \State $\mathcal{E}^+_i\gets\text{MVE}\left(\mathcal{X}^+_i\right), \ \mathcal{E}^-_i\gets\text{MVE}\left(\mathcal{Y}^-_i\right)$ \Comment{CP \eqref{Eq:MVE}}
                \State $\mathcal{E}^+\gets\{\mathcal{E}^+_i\}, \ \mathcal{E}^-\gets\{\mathcal{E}^-_i\}$
                \State $\mathcal{X}\gets\mathcal{X}\setminus \mathcal{X}^+_i, \ \mathcal{Y}\gets\mathcal{Y}\setminus \mathcal{Y}^-_i$
                \State $\mathcal{E}_X,\mathcal{E}_Y\gets$ MVEs of $\mathcal{X}$ and $\mathcal{Y}$
            \EndIf
            \State $i\gets i+1$
            
        \EndWhile
        \If{$|\mathcal{X}|(\text{or }|\mathcal{Y}|) \geq n$}
            \State $\mathcal{E}^+\left(\mathcal{E}^-\right)\gets$ MVE of $\mathcal{X}(\mathcal{Y})$
        \Else
            \State Consider points in $\mathcal{X}(\mathcal{Y})$ as individual ellipsoids
        \EndIf
        
    \end{algorithmic}
\end{algorithm}

\begin{algorithm}
    \caption{RCH Algorithm}\label{alg:RCH}
    \begin{flushleft}
        \textbf{Input}: Two datasets $\mathcal{X}$ and $\mathcal{Y}$ with different labels \\
        \textbf{Output}: $\mathcal{X}^+,\mathcal{Y}^-$ and their respective MVEs $\mathcal{E}_{X^+},\mathcal{E}_{Y^-}$
    \end{flushleft}

    \begin{algorithmic}[1]
        \State $D\gets1/\min(|\mathcal{X}|,|\mathcal{Y}|)$
        \State $\mathbf{X},\mathbf{Y}\gets \ \text{Matrices with elements of }\mathcal{X} \ \text{and } \mathcal{Y}$, respectively
        \State $\mathcal{E}_X,\mathcal{E}_Y\gets \ \text{MVEs of } \mathcal{X} \ \text{and } \mathcal{Y}$, respectively \Comment{Solve CP~\eqref{Eq:MVE} }
        \If{$\mathcal{E}_X\cap\mathcal{E}_Y=\emptyset$} \Comment{Solve CP in \cite[Sec.~15]{pope2008algorithms}}
            \State Datasets $\mathcal{X}$ and $\mathcal{Y}$ are disjoint sets
            \State $\mathcal{X}^+\gets\mathcal{X}, \ \mathcal{X}^-=\emptyset$
            \State $\mathcal{Y}^-\gets\mathcal{Y}, \ \mathcal{Y}^+=\emptyset$
            \BREAK
        \Else    
            \State $\mathbf{u},\mathbf{v}\gets$ Solution to CP~\eqref{Eq:RCH} using $D,\mathbf{X}$, and $\mathbf{Y}$
            \State $\mathbf{c}\gets\mathbf{X}^T\mathbf{u}, \ \mathbf{d}\gets\mathbf{Y}^T\mathbf{v}$ \Comment{Points in the CHs of $\mathcal{X}$ and $\mathcal{Y}$}
            \State $\mathbf{w}=\mathbf{c}-\mathbf{d}, \ \alpha=\mathbf{c}^T\mathbf{w}, \ \beta = \mathbf{d}^T\mathbf{w}$ \Comment{Parameters of hyperplanes through $\mathbf{c},\mathbf{d}$}             
            \State $\mathcal{X}^+\gets \{x_i: x_i'\mathbf{w}\geq\alpha\}, \ i=1,\cdots, N$ \Comment{Points of $\mathcal{X}$ that lie on one side of the hyperplane}
            \State $\mathcal{Y}^-\gets \{y_j: y_j'\mathbf{w}\leq\beta\}, \ j=1,\cdots, M$ \Comment{Points of $\mathcal{Y}$ that lie on the other side of the hyperplane}
            \State $\mathcal{E}_{X^+}\gets$ MVE of $\mathcal{X}^+, \ \mathcal{E}_{Y^-}\gets$ MVE of $\mathcal{Y}^-$
        \EndIf
    \end{algorithmic}
\end{algorithm}

\subsubsection{First Partition}

To find the first ellipsoidal partitions, it is assumed that the sets $\mathcal{X},\mathcal{Y}$ satisfy the I-D inequality, that is, they contain more points than the dimension of the feature space. Now, with $D=\frac{1}{\min(N,M)}$, where $N=|\mathcal{X}|$ and $M=|\mathcal{Y}|$, the RCH algorithm, Alg.~\ref{alg:RCH}, finds the SVs $\mathbf{c}$ and $\mathbf{d}$ (line 12 in Alg.~\ref{alg:RCH}) as well as the hyperplanes passing through these SVs; these are of the form $z'\mathbf{w}=\alpha$ and $z'\mathbf{w}=\beta$, respectively. Consider the simple case when these SVs are some distance apart. Now, the points in $\mathcal{X},\mathcal{Y}$ that can be separated using these hyperplanes are identified (lines 13-14 in Alg.~\ref{alg:RCH}); these points form the sets $\mathcal{X}^+,\mathcal{Y}^-$, which are covered by the MVEs $\mathcal{E}_{X^+}$ and $\mathcal{E}_{Y^-}$, respectively.

Note that though the choice of $D=\frac{1}{\min(N,M)}$ does lead to RCHs of at least one set, say $\mathcal{X}$ if $N>M$, or both $\mathcal{X}$ and $\mathcal{Y}$ if $N=M$, the RCHs, and corresponding MVEs, of either set can still contain a few points of the other; this result is dependent on the nature of the dataset. The references \cite{bennett2000duality,Goodrich2009} only indicate that the CHs reduce for $D<1$, but present no results that for a specific value of $D$, the RCHs contain points of only one label. 

Suppose $\mathcal{E}_{X^+}$ and $\mathcal{E}_{Y^-}$ contain fewer than $n_{\text{Imp}}\geq0$ points of the other label, then, $\mathcal{E}_{X^+}$ and $\mathcal{E}_{Y^-}$ become the first ellipsoidal partitions. On the other hand, if one of them fails this condition, say $\mathcal{E}_{X^+}$, then, the RCH algorithm is reapplied on the sets $\mathcal{X}^+$ and $\mathcal{Y}$ with the choice $D=\frac{1}{\min(|\mathcal{X}^+|,|\mathcal{Y}|)}$ (lines 13-15 in Alg.~\ref{alg:EllipPart}). Note that $|\mathcal{X}^+|<|\mathcal{X}|$ if the dataset is not linearly separable, while $|\mathcal{X}^+|=|\mathcal{X}|$ otherwise. Thus, in the former case, the RCH algorithm is repeatedly applied on the sets $\mathcal{X}^+,\mathcal{Y}$, by choosing $D=\frac{1}{\min(|\mathcal{X}^+|,|\mathcal{Y}|)}$ in every iteration, such that $\mathcal{X}^+$ contains a few points of $\mathcal{X}$ and no more than $n_{\text{Imp}}$ points of $\mathcal{Y}$ - at the end of these iterations emerges the the reduced CH, and MVE, of $\mathcal{X}$. The same procedure is applied to $\mathcal{Y}^-$, if needed. Fig.~\ref{fig:RCH_it1} shows these first partitions of the 2-D synthetic dataset shown in Fig.~\ref{fig:Init_CH_Ellp}; the value of $n_{\text{Imp}}=10$ is used. Since the CHs reduce in each iteration, the MVEs in the final iteration are expected to be non-intersecting. 

\textbf{Remarks}: \textit{i}. The $n_{\text{Imp}}\geq0$ points that are permitted may be interpreted as being ``impure'' points, borrowing the terminology of DT classifiers. In the ideal case, $n_{\text{Imp}}=0$ should be the only choice exercised by the user. \textit{ii}. It can happen, for some value of $n_{\text{Imp}}$, that the set $\mathcal{X}^+$, or $\mathcal{Y}^-$, violates the I-D inequality, for example, $\left(|\mathcal{X}^+|+n_{\text{Imp}}\right)<n$. In this case, since an MVE cannot be found that covers these points, these are retained as isolated points. This situation can occur if the dataset has heavy overlap with only a few points of either label that can be ``separated''. \textit{iii}. Suppose $N=M$, that is, the datasets are balanced, then, with this choice of $D$, the solutions $\mathbf{c}$ and $\mathbf{d}$ are the centroids of the sets $\mathcal{X}$ and $\mathcal{Y}$, respectively; this result is also demonstrated in \cite{Goodrich2009}. On the other hand, when $N>M$, the solution $\mathbf{d}$ is the centroid of $\mathcal{Y}$, while the solution $\mathbf{c}$ leads to an RCH of $\mathcal{X}$ that has shrunk non-uniformly towards the centroid of $\mathcal{X}$, \cite{Goodrich2009}. \textit{iv}. The RCH algorithm aims to find $\mathbf{c}$ and $\mathbf{d}$ that are separated from each other by a non-zero distance. If this fails, as we show for the 2-D XOR problem, then, infinite number of separating hyperplanes that contain $\mathbf{c}$ and $\mathbf{d}$ can be found. In such cases, alternative interpretations of the partitions are provided.

\subsubsection{Further Partitions}

Having obtained the first partitions, the points contained in $\mathcal{X}^+$ and $\mathcal{Y}^-$ are removed from the dataset - lending the sequential component of SEP-C - and the process is continued for the smaller sets $\mathcal{X}\setminus \mathcal{X}^+$ and $\mathcal{Y}\setminus \mathcal{Y}^-$ (line 22 in Alg.~\ref{alg:EllipPart}). Note that if the points are linearly separable, then $\mathcal{X}\setminus \mathcal{X}^+=\emptyset$ and $\mathcal{Y}\setminus \mathcal{Y}^-=\emptyset$; now SEP-C terminates since both sets violate the I-D inequality. On the other hand, if the dataset has some overlap, then, partitioning happens for the points that lie within or close to the region of overlap. For example, as shown in Fig.~\ref{Fig:EPart_Ov}, for the points marked in blue, the first partition is the left-most ellipse and the subsequent elliptical partitions are found from left to right; for the points marked in red, the opposite is true. As is expected, since these ellipses focus on the points in the overlap region, these can be expected to overlap as well. 


\subsubsection{Terminal Partitions}

SEP-C is guaranteed to terminate in a finite number of iterations. The conditions for termination are when either all points have been partitioned and the RCH algorithm fails (line 8 in Alg.~\ref{alg:EllipPart}) to find even further RCHs for each set or the points of one set violate the I-D inequality. Since one of these conditions is bound to hold true, simply because the training set starts with a finite size, which keeps reducing in every iteration, SEP-C is bound to terminate as well. The number of iterations is again dependent on the dataset; as we show in the Results section, different training sets require different number of iterations. Note that when SEP-C terminates, there could be a few points of each label that cannot be covered by an MVE, in this case, these ``leftover'' points are treated individually (line 30 in Alg.~\ref{alg:EllipPart}). In the case of imbalanced datasets, where the datapoints of the minority class are fully partitioned, but some from the majority class remain to be partitioned, lines 27-31 in Alg.~\ref{alg:EllipPart}, ensure that the remainder of the majority class are covered by an MVE. 


\subsection{Features of the Algorithm}

The main feature that is immediately evident is that since partitioning happens sequentially - essentially on a different subset of the training set in each iteration - SEP-C finds multiple hyperplanes. This feature is unlike traditional SVM classifiers and their variants, where a single hyperplane is formulated by solving a CP. As will be shown in the Results section, especially for the non-linearly separable datasets, these multiple hyperplanes can be interpreted as being a piece-wise linear approximation of the non-linear separating hyperplane. Further, SEP-C also adopts an iterative approach in implementing the RCH algorithm, rather than a single choice to find these RCHs. Multiple MVEs are found, rather than multiple convex hulls, since, in higher dimensions, computing convex hulls can be computationally expensive, \cite{Chazelle1993}. Each of the MVEs, of the form $\mathcal{E}=\{\mathbf{v} \ | \ \norm{\mathbf{A}\mathbf{v}+\mathbf{b}}\leq 1\}$, that partition the dataset can be interpreted as one that models the dataset as a local multivariate normal distribution - the centre of the ellipsoid, $\mathbf{b}$, is the mean and the matrix, $\mathbf{A}$, is the covariance matrix. While the GMM approach similarly models a dataset as being composed of several normal distributions, the number of mixtures is a user-defined input in the GMM, while in our approach, the number of such mixtures is a result of the nature of the dataset. 

SEP-C uses the user-defined parameter $n_{\text{Imp}}$ to find the MVEs in each iteration. As we discuss next, this aspect makes the rules of classification more interpretable, in comparison with the use of regularisation parameters in SVMs, where the width of the margin is varied to allow for classification errors. The conscious inclusion of $n_{\text{Imp}}$ number of misclassifications ensures that the MVEs obtained in any iteration do not become ``small'' and also limits the number of iterations that the algorithm uses. The $n_{\text{Imp}}>0$ number of ``impure'' points could be owing to outlier datapoints of one label. As we discuss in the section on classification rules, the inclusion of these ``impure'' points considers the influence of such points explicitly in predicting the label of a new datapoint. It is highlighted that the algorithm does not need for outliers to be removed prior to its application and further, their locations in one or more MVEs are now known.

The ellipsoidal partitions obtained using Alg.~\ref{alg:EllipPart} are used to state the rules of classification of a new test point. Briefly, the test datapoint is assigned a label based on which ellipsoid contains this point and the number of training datapoints of either label that are also contained within that ellipsoid. This feature of the algorithm is presented next. 

\subsection{Rules of Classification}\label{Sec:RulesClassify}

SEP-C is a type of sequential-covering algorithm, \cite{MitchellMLBook}. In such algorithms, a single rule is learnt that can distinguish a subset of positive and negative examples. The datapoints covered by this single rule are then removed from the dataset and new rules are learnt on the remaining data. Each rule is expected to have high accuracy but is not expected to cover all examples from the training set. Accuracy is again measured in terms of some performance measure on the training examples, for example, the number of correct predictions.

In SEP-C, the rules of classification are based on the MVEs found by partitioning; we have designed SEP-C for binary classification. Let the partitioned ellipsoids $\mathcal{E}_{k}^+, \ k=1,2,\cdots$, be those that correspond to points with label $L_{+1}$ and $\mathcal{E}_{j}^-, \ j=1,2,\cdots$, correspond to label $L_{-1}$. The performance measure that forms the basis of partitioning is by determining the MVEs that mostly contain points with the same label and at most $n_{\text{Imp}}$ number of the other label; thus, high accuracy in partitioning can be expected as each partition correctly classifies those examples contained within it. Note that each of the MVEs, for example, as shown in Fig.~\ref{Fig:EPart_Ov}, cover only a part of the dataset.

With this approach, rules of classification of a new test point are considered for the following cases: When the test point is 
\begin{enumerate}[label=\textbf{Case}~\arabic*, leftmargin=*]
    \item contained within only \textit{one} of the MVEs $\mathcal{E}_{k}^+$ or $\mathcal{E}_{j}^-$;
    \item contained in the region of intersection of one or more MVEs $\mathcal{E}_{k}^+$ and $\mathcal{E}_{j}^-$; and
    \item \textit{not} contained in any of them.
\end{enumerate}
As we show, \textbf{Cases}~1 and 2 may be considered when one or more rules covers the test point and \textbf{Case}.~3 for the case when the test point is not covered. In standard sequential-covering algorithms, a test point that is not covered by any of the rules is usually assigned the default class. To avoid resorting to this situation, we propose a rule under \textbf{Case}~3.

\textbf{Case}~1: Let $z_1$ denote test datapoint that is contained in \textit{only one} of the ellipsoids. For this point, its label is assigned according to the rule
\begin{equation}\label{z1labelSE}
    R_1\text{: Label of }z_1=\begin{cases} L_{+1} & \text{if } z_1\in\mathcal{E}_{k}^+ \\
    L_{-1} & \text{if } z_1\in\mathcal{E}_{j}^-
    \end{cases}
\end{equation}
This assignment of the label to $z_1$ is natural as the MVEs are partitioned with high accuracy, or its control based on $n_{\text{Imp}}$. Suppose $z_1\in\mathcal{E}_{k}^+$, but $\mathcal{E}_{k}^+$ intersects with one or more MVEs from either $\mathcal{E}^+$ or $\mathcal{E}^-$. In such cases, it may happen that $z_1$, now the label $L_{+1}$, is ``close'' to points with label $L_{-1}$ leading to a possible misclassification; proximity between points can be measured, say using Euclidean distance or similar measure. Although misclassification can be identified only \textit{a posteriori}, we propose a trust-scoring mechanism by which the possibility of misclassification can be highlighted; this is considered explicity in the next section.

\textbf{Case}~2: Let $z_2$ denote the test datapoint that is contained in the region of intersection of two or more MVEs, $\mathcal{E}_v, \ v=k,j$; denote $R_{\text{Int}}=\bigcap_{v=k,j}{\mathcal{E}_v}$ as the region of intersection. This case can be further classified into: if $R_{\text{Int}}$ contains \textit{i}. a non-zero number of points with both labels from the training set and \textit{ii}. only $z_2$. For the first case, the rule of classification is
\begin{equation}\label{z2labelR2a}
    R_{2a}\text{: Label of }z_2=\begin{cases} L_{+1} & \text{if } |\mathcal{X}\in R_{\text{Int}}|> |\mathcal{Y}\in R_{\text{Int}}| \\
    L_{-1} & \text{if } |\mathcal{X}\in R_{\text{Int}}|<|\mathcal{Y}\in R_{\text{Int}}|
    \end{cases},
\end{equation}
and for the second,
\begin{equation}\label{z2labelR2b}
    R_{2b}\text{: Label of }z_2 =\begin{cases} L_{+1} & \text{if } |\mathcal{X}\in \mathcal{E}_U|> |\mathcal{Y}\in \mathcal{E}_U| \\
    L_{-1} & \text{if } |\mathcal{X}\in \mathcal{E}_U|< |\mathcal{Y}\in \mathcal{E}_U|,
    \end{cases}
\end{equation}
where $\mathcal{E}_U = \bigcup_{v=k,j}{\mathcal{E}_v}$. 

The case when the strict inequality does not hold, either in \eqref{z2labelR2a} or \eqref{z2labelR2b}, is considered in Sec.~\ref{Sec:TrustScore}.

\textbf{Case} 3: Let $z_3$ be the test point that satisfies this case. Now, we find the MVE, say $\mathcal{E}_w$, that is closest to this test, expand $\mathcal{E}_w$ to cover this point, and then apply rule $R_1$, or one of rules $R_{2a}$ or $R_{2b}$, to assign the label to this point. The point $z_3$ now lies on the boundary of $\mathcal{E}_w$. Note that determining the distance between a point and an ellipsoid is also a CP, \cite{pope2008algorithms}. It can happen that $z_3$ is equidistant from more than one MVE. Now, each of these MVEs is expanded to cover this point and rules $R_1$ or $R_2$ are applied. 

It can be observed that the rules proposed are mutually exclusive, for example, a test point cannot lie in a single MVE and simultaneously lie in a region of overlap of several MVEs. The same holds for the other rules as well. As argued in \cite{Huysmans2008}, for mutually exclusive rules, they do not need to be ordered in any way or their combination be checked prior to assigning a label to a test point. Since the rules $R_1$ and $R_{2}$ are based on the partitioned MVEs, we subscribe to the statement in \cite{Huysmans2008} that ``each rule in itself represents a part of the input space and can be considered as a nugget of knowledge'' and consider each MVE as a ``nugget of knowledge''. 

\subsection{Trust in Classification}\label{Sec:TrustScore}

The rules presented do not provide any additional measure of trust or confidence in the prediction; such a trust calculation is also missing in the cited references that propose rules for SVMs. For example, let a test point get a label according to Rule~$R_1$, \eqref{z1labelSE}. In SEP-C, we ask the additional question: ``Can this prediction be trusted?'' This question is important since, even though the MVEs found in each iteration do not overlap with each other, when all MVEs are considered, some may overlap with each other, such as shown in Fig.~\ref{Fig:EPart_Ov}. Such overlaps typically occur close to the main region of overlap of the dataset itself; see the MVE marked in green in Fig.~\ref{Fig:Init_Ellps}. In addition, SEP-C permits the inclusion of $n_{\text{Imp}}\geq0$ number of ``impure'' points.  Thus, the question posed on trust helps answer if the prediction is possibly ``more correct'' if an MVE, or the overlapping region of MVEs, contains more points with one label than the other along with impure points. 

We calculate trust in classification of a test point as a conditional, or posterior, probability determined from Bayes' theorem, in the form  
\begin{align}\label{Eq:PostProbLabel}
    P(L_{+1}|\mathcal{E}_v) &= \frac{P(\mathcal{E}_v|L_{+1})P(L_{+1})}{P(\mathcal{E}_v)} \nonumber \\
    &= \frac{P(\mathcal{E}_v|L_{+1})P(L_{+1})}{P(\mathcal{E}_v|L_{+1})P(L_{+1}) + P(\mathcal{E}_v|L_{-1})P(L_{-1})},
\end{align}
where, $\mathcal{E}_v$ is the MVE that contains the test point and $P(L_{+1})$ (or $P(L_{-1})$) is the prior probability of the point having label $L_{+1}$ (or $L_{-1}$). By considering $\mathcal{E}_v$ as either the intersection of two or more MVEs, or even their union, the formula \eqref{Eq:PostProbLabel} is applicable for all the stated rules. 

In \eqref{Eq:PostProbLabel}, the prior probabilities $P(L_{+1})$ and $P(L_{-1})$ are determined based on the number of occurrences of points of these label in the training set and the predicted label of the test point. Thus,
\begin{equation}\label{Eq:PriorProbLp1Lm1}
    P(L_{+1}) = \frac{N+1}{N+M+1}, \ P(L_{-1}) = 1-P(L_{+1}), 
\end{equation}
where, $N$ and $M$ are the number of points in the training dataset with labels $L_{+1}$ and $L_{-1}$, respectively. In addition, if $n$ number of points with label $L_{+1}$ and $m$ number of points with label $L_{-1}$ are contained within $\mathcal{E}_v$, then, the conditional probabilities
\begin{equation}\label{Eq:ProbLp1LM1Ev}
    P(\mathcal{E}_v|L_{+1}) = \frac{n+1}{n+m+1}, \ P(\mathcal{E}_v|L_{-1}) = 1-P(\mathcal{E}_v|L_{+1}).
\end{equation}
Based on these values, the trust in prediction can be computed as a posterior probability. 

We demonstrate this calculation for a few simple cases. Let the test point, $z$, be contained in an MVE, $\mathcal{E}_k^+$. Further, let $\mathcal{E}_k^+$ be such that it does not overlap with any other MVE and also contains $n_{\text{Imp}}$ number of points with label $L_{-1}$; now, Rule~$R_1$ is triggered and $z$ gets the label $L_{+1}$. From \eqref{Eq:PostProbLabel}, the trust in this prediction is given by
\begin{equation}\label{Eq:PostProbRule1Simp}
    P(L_{+1}|\mathcal{E}_k^+) = \frac{1}{1+\dfrac{n_{\text{Imp}}M}{(n+1)(N+1)}}.
\end{equation}
We now consider two cases: \textit{i}. a balanced dataset, where the ratio $\frac{M}{N+1}\approx 1$ and \textit{ii}. an imbalanced dataset, where $M\ll N$. Now, in the balanced dataset case,
\begin{equation}\label{Eq:PostProbRule1SimpBD}
    P(L_{+1}|\mathcal{E}_k^+) \approx \frac{1}{1+\dfrac{n_{\text{Imp}}}{(n+1)}}.
\end{equation}
Clearly, if $n_{\text{Imp}}\ll(n+1)$, the trust in prediction of $L_{+1}$ ($L_{-1}$) is high (low) - close to 1 (0). 

Consider the case of an imbalanced dataset, where points with label $L_{+1}$ form the majority class; thus, in \eqref{Eq:PostProbRule1Simp}, the ratio $\frac{M}{N}\ll1$, again indicating a high trust for label $L_{+1}$, if $z$ lies in $\mathcal{E}_k^+$. This is a natural result, since the dataset is skewed in favour of the majority class. This result may indicate that it may not be possible at all to label a point of the minority class, with label $L_{-1}$, with high trust. However, since SEP-C tries to find an MVE, say $\mathcal{E}_j^-$, that contains points of \textit{only} the minority class, that is $n\approx0$, then \eqref{Eq:PostProbRule1Simp} leads to the trust $P(L_{+1}|\mathcal{E}_j^-)\approx0\Rightarrow P(L_{-1}|\mathcal{E}_j^-)\approx1$, indicating a high trust in label $L_{-1}$. 

For datasets with a large region of overlap, SEP-C may find very few ellipsoidal partitions, even for a high value of $n_{\text{Imp}}$. In such cases, it is clear, from \eqref{Eq:PostProbRule1Simp}, that the trust will be close to 50\% for the prediction, which we recommend as a ``low'' trust score. Another way to interpret why this may be considered low is by computing the odds ratio $P(L_{+1}|\mathcal{E}_k^+)/P(L_{-1}|\mathcal{E}_k^+)=1$, that is, the odds are nearly the same that the test point can have either label. For the case when the odds ratio is close to 1, it is evident that \textit{the classifier should abstain from assigning a label} to the test point and that user intervention is required. This result is equivalent to when the test point lies exactly on the separating hyperplane in classical SVMs. Instead of resorting to applying the label of the default class to such a test point, which may hold in imbalanced datasets, using the language of \cite{Fumera2002}, we prefer that the classifier ``rejects'' classifying this test point. It should be highlighted that the formulation of the SVM with reject option presented in \cite{Fumera2002} is based on limiting the values of the coefficients by which the support vectors are found and thus shares a similar idea as that of the RCH, which we adopt. Note that their formulation also does not use a convex function, unlike our approach, where only CPs are solved at every step.

Following this approach, a trust score can be calculated if the test point triggers Rules~$R_{2a,2b}$. This approach is also valid if, in \eqref{z2labelR2a} and \eqref{z2labelR2b}, $|\mathcal{X}\in R_{\text{Int}}|=|\mathcal{Y}\in R_{\text{Int}}|$ or $|\mathcal{X}\in \mathcal{E}_U|= |\mathcal{Y}\in \mathcal{E}_U|$, respectively. When we apply SEP-C on datasets, we compute trust scores for those regions in which the test points lie. Note that the number of such regions - disjoint MVEs or overlapping MVEs - are finitely many. 

\textbf{Interpreting Rules and Trust Scores}: In the literature, several metrics have been proposed to analyse the quality of rules extracted from classifiers; see \cite{martens2009} and \cite{Barakat2007} for a description and the methods of evaluation. A key requirement of a rule is that it is comprehensible. For example, \textit{If-Then} rules are crisp, but the number of conditions required in such rules may reduce comprehensibility. In SEP-C, the rules of classification and the trust scores rely on the properties of the ellipsoid(s) that contain the test point. Let Rule~$R_1$ be triggered and a test point gets the label $L_{+1}$ as it is contained within the MVE $\mathcal{E}_k^+$. Let also the trust score formula, \eqref{Eq:PostProbRule1Simp}, yield a high trust for the label $L_{+1}$. This result can be interpreted as saying that since the test point has characteristics that are similar to a greater number of points with label $L_{+1}$ in $\mathcal{E}_k^+$, higher are the odds that the test point too has the label $L_{+1}$; the converse also holds. A similar interpretation can be made when one of Rules~$R_{2a}$ or $R_{2b}$ is triggered, that is, when the test point is contained in the region of overlap or in the union of the intersecting MVEs. These points indicate that the trust score may also be thought of as a nonconformity measure, \cite{shafer2008tutorial}, which is a ``real-valued function $A(B,z)$ that measures how different an example $z$ is from the examples in a bag $B$''. The trust score, especially when Rule~$R_{2a}$ or $R_{2b}$ is triggered, can also be compared with the concept of the overlapping coefficient (OVL), that is used to measure the similarity between probability distributions, \cite{Inman1989,SCHMID2006,Kalai2010}. Given two probability density functions $f$ and $g$, a test point, $z$, and the decision rule that $z$ belongs to $f$, iff $f(z)\geq g(z)$, and to $g$, otherwise; in \cite{SCHMID2006}, OVL is considered as the sum of two error probabilities. Note that $\text{OVL }=1$ if $f$ and $g$ match and is 0, if they are disjoint. Thus, by measuring the overlap of the ellipsoids, which are Gaussians, using the formulae presented in \cite{Inman1989,SCHMID2006,Kalai2010}, a bound on the error probability of assigning a particular label to the test point can be determined. 

\subsection{Time Complexity Analysis}\label{Sec:TimeComplexity} 

The main components of SEP-C are functions that determine the RCHs and the MVEs. Since the RCH algorithm is a dual of the SVM, \cite{bennett2000duality}, its time complexity is the same as that of the SVM, which is $\mathcal{O}(n^3)$, where $n$ is the training set size, \cite{SVMSolvers}. The solvers in the \textit{cvx-opt} modules used to find MVEs are based on state-of-the-art Primal-Dual algorithm, which is a type of Interior Point Method (IPM), which in turn is based on Newton's method applied to modified KKT conditions. In \cite{cvx}, Primal-Dual method is shown to be faster than the Barrier method - analysed in detail in \cite{cvx} - which is another IPM. The complexity of these algorithms depends on the complexity of the Newton steps required to find the optimal solution. For a fixed duality gap, in the Barrier method, the Newton steps grow as $\sqrt{m}$, where $m$ is the number of inequality constraints in the CP, hence, the Barrier method has complexity of order $\mathcal{O}(\sqrt{m)}$. In SEP-C, both $n$ and $m$ reduce in every iteration; note that the number of iterations is dependent on the dataset. 

\section{Results}\label{section:Results}

SEP-C is demonstrated on various types of 2-D synthetic datasets as well as the \textit{Adult}, \textit{Wisconsin Breast Cancer-Diagnostic (WDBC)}, and \textit{Vertebral Column (VC)} Datasets, \cite{Dua:2019}. The CP solver \textit{cvx-opt module SDPT3}, \cite{cvx}, is used in MATLAB 2021b to obtain all results; other solvers such as MOSEK and SeDuMi yielded similar results. As a first step, in most cases, the overlap between the MVEs of points of both labels is examined. This is performed by computing the overlap ratio (OVR) as the ratio of volume of the MVE of overlapping region to the volume of the MVE of either class; note that the volume of an ellipsoid, $\mathcal{E} = \{z \ | \ \norm{\mathbf{A}z+\mathbf{b}}\leq 1\}$, is proportional to $\sqrt{det(\mathbf{A}^{-1})}$, \cite[Chap.~3]{doi:10.1137/1.9781611970777.ch3}. Thus, an OVR close to 1 indicates fully overlapping classes, also implying potentially large misclassifications, and a value closer to 0 indicates little or no overlap, implying linear separability. 

As we show, the selected datasets display different overlapping properties. For example, the WDBC dataset has little overlap; the Adult dataset has significant overlap of points of both labels; and in the VC dataset, nearly all points of one label - the minority class - are in the overlap region, while only a few points of the majority class are in this region. Choosing the MVE approach to cover points of a label, prior to training and classification, allows for such data landscapes to be drawn.

\subsection{2-D datasets}

\begin{figure*}[tpbh!]
    \centering
    \begin{subfigure}[b]{0.33\textwidth}
    \centering
        \includegraphics[scale=0.4]{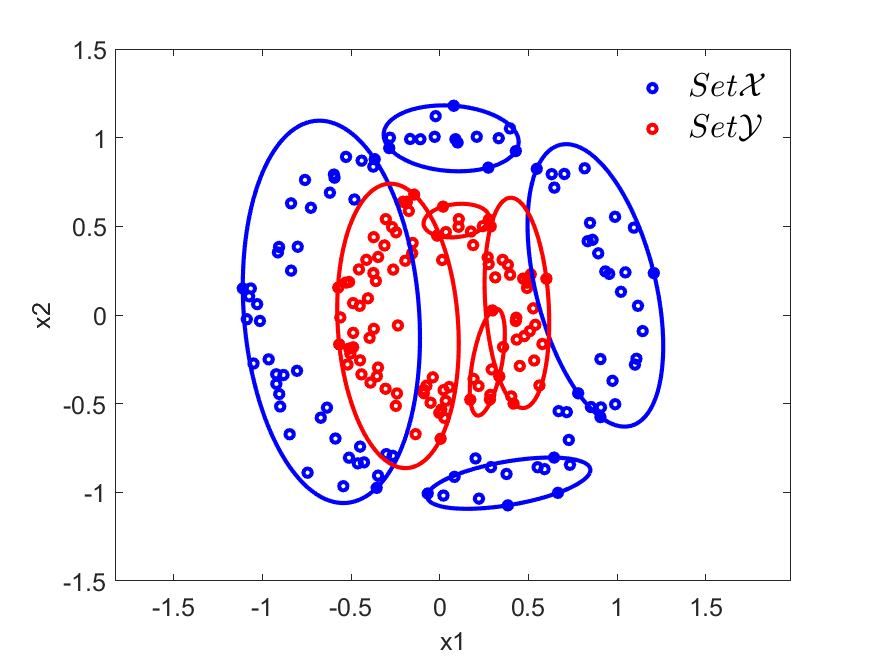}
        \caption{\label{Fig:Ellps_circle}}
    \end{subfigure}
    \begin{subfigure}[b]{0.33\textwidth}
    \centering
        \includegraphics[scale=0.4]{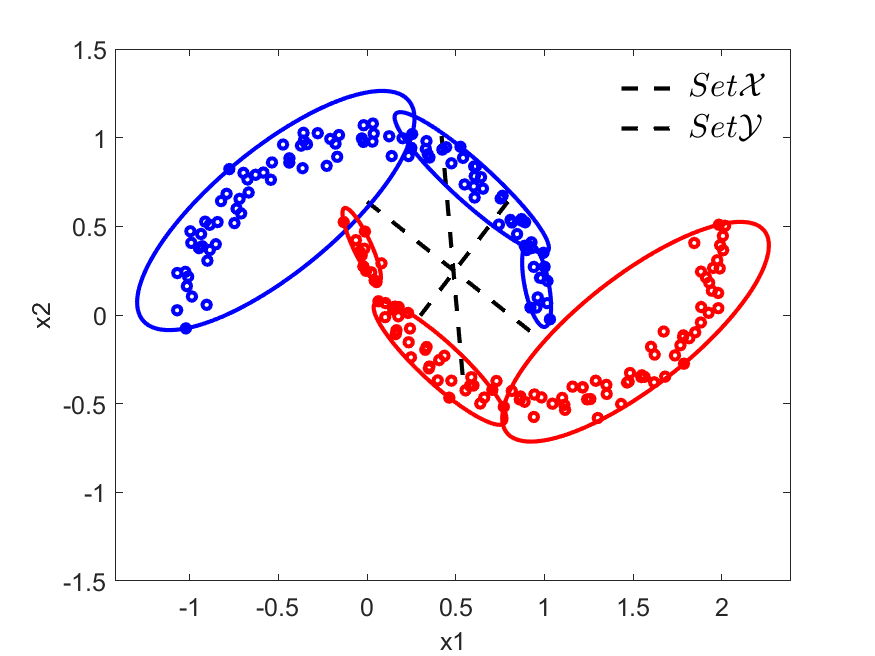}
        \caption{\label{Fig:Ellps_moon}}
    \end{subfigure}
    \begin{subfigure}[b]{0.33\textwidth}
    \centering
        \includegraphics[scale=0.4]{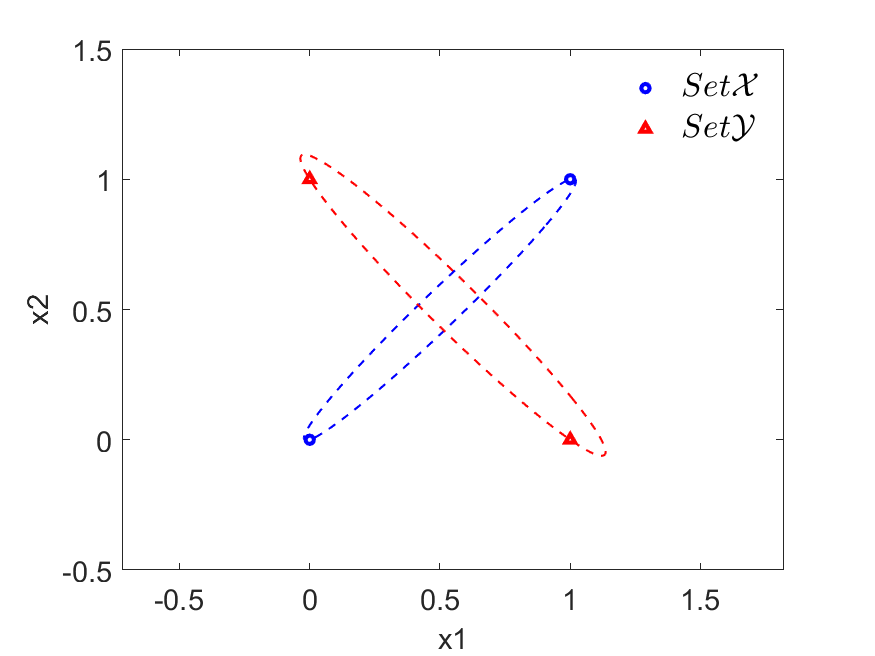}
        \caption{\label{Fig:XOR2D}}
    \end{subfigure}
    \caption{2-D Synthetic Datasets. \ref{Fig:Ellps_circle}: Circles; \ref{Fig:Ellps_moon}: Moons; and \ref{Fig:XOR2D}: XOR}
    \label{fig:circle_moon_xor}
\end{figure*}

The results of applying SEP-C on 2-D datasets that \textit{cannot} be separated by a single linear hyperplane are shown in Fig.~\ref{fig:circle_moon_xor}; these are the Circles\footnote{\url{https://scikit-learn.org/stable/modules/generated/sklearn.datasets.make_circles.html}}, Moons\footnote{\url{https://scikit-learn.org/stable/modules/generated/sklearn.datasets.make_moons.html}}, and the dataset formed by the outputs of a 2-D XOR gate. As can be seen in Figs.~\ref{Fig:Ellps_circle} and \ref{Fig:Ellps_moon}, where, the points marked in blue belong to one label and the ones in red have the other label, SEP-C is able to partition points - after 4 iterations for Circles and 3 for Moons - of either label into their respective ellipsoids. To obtain these results, the number of permitted impure points is set to $n_{\text{Imp}}=5$ and $n_{\text{Imp}}=2$ respectively. It is known that the Circles dataset can be partitioned using a linear hyperplane by transforming the features using a non-linear function and both the Circles and the Moons datasets can be partitioned using a single non-linear hyperplane defined, say using Radial Basis Functions. On the other hand, SEP-C can be applied directly without the use of any such transform, non-linear hyperplane, or the ``kernel trick''. Thus, SEP-C obviates the need of EDA, which becomes essential in determining the transform that aids in obtaining linear separability. The muliple linear hyperplanes found in each iteration of SEP-C are marked in black in Fig.~\ref{Fig:Ellps_moon}; similarly, multiple such linear hyperplanes are also found in the Circles dataset.

Consider the XOR dataset shown in Fig.~\ref{Fig:XOR2D}. Following Alg.~\ref{alg:EllipPart}, SEP-C does \textit{not} perform the first partition (the main \texttt{while} loop in line 7) as the number of points exactly equals the dimension of the feature space. Even if the inequality in line 7 in Alg.~\ref{alg:EllipPart} is relaxed, that is, $|\mathcal{X}|\geq n$, the RCH algorithm, which is called in line 8, cannot find non-overlapping MVEs, leading to empty sets $\mathcal{X}^+$ and $\mathcal{Y}^-$, and eventually exiting the \texttt{while} loop (line 9). This result is a consequence of Radon's theorem, \cite{Adams2020}, as non-intersecting convex hulls cannot be found for these set of points. In either case, in line 27 which is outside the main \texttt{while} loop, SEP-C finds MVEs that cover points of both sets; these MVEs necessarily have to intersect. 

\textbf{Remarks}: \textit{i}. Since in 2-D, the convex hulls formed by the points at the diagonally opposite vertices of the square are essentially straight lines, the lengths of the minor axes of the corresponding MVEs are zero, implying that the corresponding eigenvalues of the matrix $\mathbf{A}$ in the ellipsoid definition $\mathcal{E} = \{z \ | \ \norm{\mathbf{A}z+\mathbf{b}}\leq 1\}$, have magnitudes equal to infinity. In this case, the chosen solver fails to find this MVE (indeed, it may be true that no solver can find such MVEs). However, by introducing just two points, where one point is ``close'' to $(0,1)$ and the other close to $(1,1)$, the solver can find the MVEs; this procedure is adopted to plot the ellipses shown in Fig.~\ref{Fig:XOR2D}. These new points are generated randomly within a radius of magnitude 0.01 around these points. \textit{ii}. Analogous to the XOR function being ``learnt'' by an Artificial Neural Network (ANN) with a hidden layer, the two MVEs found by SEP-C can be interpreted as being the same function; see \cite{Bland1998} for challenges faced in the design of ANNs for the XOR problem.


\subsection{Adult Dataset}

The Adult (Census Income) Dataset consists of 30162 records, with 22654 records with one label, say $L_{+1}$, and 7508 records of the other; as is evident, this dataset is highly imbalanced. Moreover, calculation of the OVRs of the MVEs of the points of both labels indicates that a significant portion of the minority class ($\text{OVR}=0.6$) lies in the overlapping region, hence, the potential for misclassification is high. Each record is defined by a mix of 14 categorical (8) and continuous (6) features. The 8 categorical features in the dataset have very few but repeated discrete values. Thus, when the entire dataset is considered, SEP-C is unable to find MVEs in many iterations owing to the constant values of a few of the categorical features in some partitions. In addition, for some combinations of the categorical features, there are \textit{no} datapoints consisting of the continuous features at all, hinting at a sparse dataset with very few informative features. Thus, the dataset is stratified and the most populated stratum is selected as the training set. As discussed in \cite{ZHAO2019416}, the approach of stratified sampling is a systematic and cost-effective way of shrinking the dataset size while preserving its properties; see also \cite{YE2013769,JING20153688} on the use of stratified sampling of features for high-dimensional data.

The selected stratum had the following values for the categorical features: $\texttt{workclass}=2$, $\texttt{education}=11$, $\texttt{education}=11$, $\texttt{marital-status}=2$, $\texttt{occupation}=2$, $\texttt{relationship}=0$, $\texttt{race}=4$, $\texttt{sex}=1$, and $\texttt{native-country}=38$. The training set from this stratum has 547 records with label $L_{+1}$ and 256 with label $L_{-1}$. Even in this stratum, one of the continuous features has a constant value for all records, thus lending it a categorical nature; this feature too is dropped, thus leading to the training set consisting of only 5 continuous variables. For this reduced dataset, the OVRs of the respective MVEs and the overlapping region is close to 50\% for both labels, again indicating the high degree of overlap. 

\begin{figure}[thpb!]
		\centering
        \resizebox{0.7\columnwidth}{!}{
		\begin{tikzpicture}
        \pgfplotsset{set layers,every tick label/.append style={font=\huge},label style={font=\huge}}
            \begin{axis}[
				scale only axis,
				xmin = 0, xmax = 820,
                ymin = 0, ymax = 120,
				xlabel = DataPoints, 
				ylabel = Values,
				ytick = {20,40,60,80,100,120},
                xtick = {0,200,400,600,810},
                axis y line*=left]
				\addplot [blue, ycomb, mark=*, mark size = 2 pt, mark options={fill=blue}, line width = 1 pt] table {Data/Feat11_Full.dat};
            \end{axis}
            \begin{axis}[
				scale only axis,
				xmin = 0, xmax = 820,
                ymin = 0, ymax = 120,
			    xtick = {0,200,400,600,810},
                ytick = \empty,
                axis y line*=right]
				\addplot [green, ycomb, mark=*, mark size = 2 pt, mark options={green}, line width = 1 pt] table {Data/Feat11_it1.dat};
			\end{axis}

        \end{tikzpicture}
    }
    \caption{Values of the feature $\texttt{capital-gain}$ for the entire dataset, marked in blue, and in the subset found by the RCH function, marked in green. As all values in green are 0, \texttt{cvx-opt} cannot find the MVE for the reduced CH.}
    \label{fig:plot_Feat11}
\end{figure}
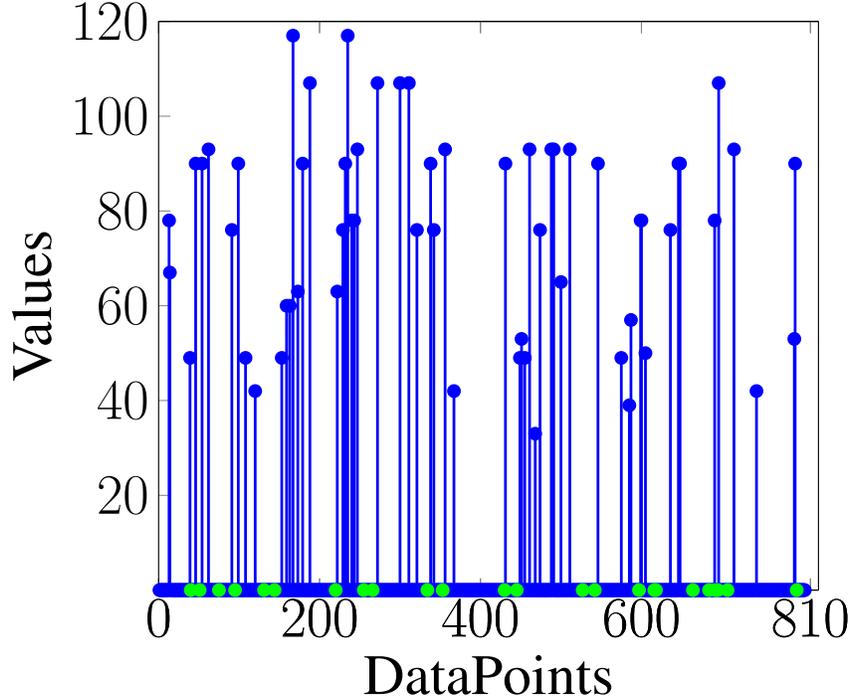

In the first iteration of SEP-C, the result of applying the RCH function led to a subset of the training data where 2 continuous features \texttt{capital-gain} and \texttt{capital-loss} have values equal to 0; Fig.~\ref{fig:plot_Feat11} shows the values of feature \texttt{capital-gain} for the entire dataset (blue) as well as the values, of magnitude 0, for the datapoints in the RCH found in the 1\textsuperscript{st} iteration (in green). Since the subset of these features at the end of the 1\textsuperscript{st} iteration are constant, the MVE could not be found, again because the corresponding semi-axes lengths of the MVE become zero. Instead of again removing these features and working with a dataset with a reduced dimension, low magnitude uniform random noise was added to these two features in the entire training set; a similar procedure is followed for the XOR problem. Now, with $n_{\text{Imp}}=10$, SEP-C led to 17 ellipsoidal partitions for each label.

The performance of SEP-C is compared with the Decision Tree Classifier (DTC) on this reduced dataset; the same 80-20 split is used for training and testing in both cases. DTC was selected to have a depth of 10 and for SEP-C, results with $n_{\text{imp}}=10$ are presented. The motivations to compare SEP-C and DTC (in addition to DTC being a well-known one) are \textit{i}. the rules of classification can be inferred and \textit{ii}. the leaf nodes in DTC are analogous to the regions formed by MVEs in SEP-C. However, differences are notable too: \textit{i}. In DTC, a test record gets the majority label, if the leaf in which it lies contains equal number of records of both labels, while in SEP-C, a label is assigned on the corresponding MVE in which it lies. Of course, in SEP-C, if the MVE contains equal number of records of both the classes, the trust score is close to 50\%; \textit{ii}. the leaf nodes in DTC are essentially polyhedra in the feature space where an edge of a polyhedron is along the direction of the feature; some of these polyhedra may be unbounded (a leaf being defined by a single inequality for one of the features, for example, $\texttt{capital-gain}> 77$ for leaf 134), while in SEP-C, the regions are bounded by the MVEs; and \textit{iii}. overfitting in SEP-C can be controlled by tweaking $n_{\text{Imp}}$ , while in DTC, overfitting is controlled by pruning, which involves tuning of more than one hyperparameters, such as depth of the tree, number of leaves, or through cost complexity pruning.

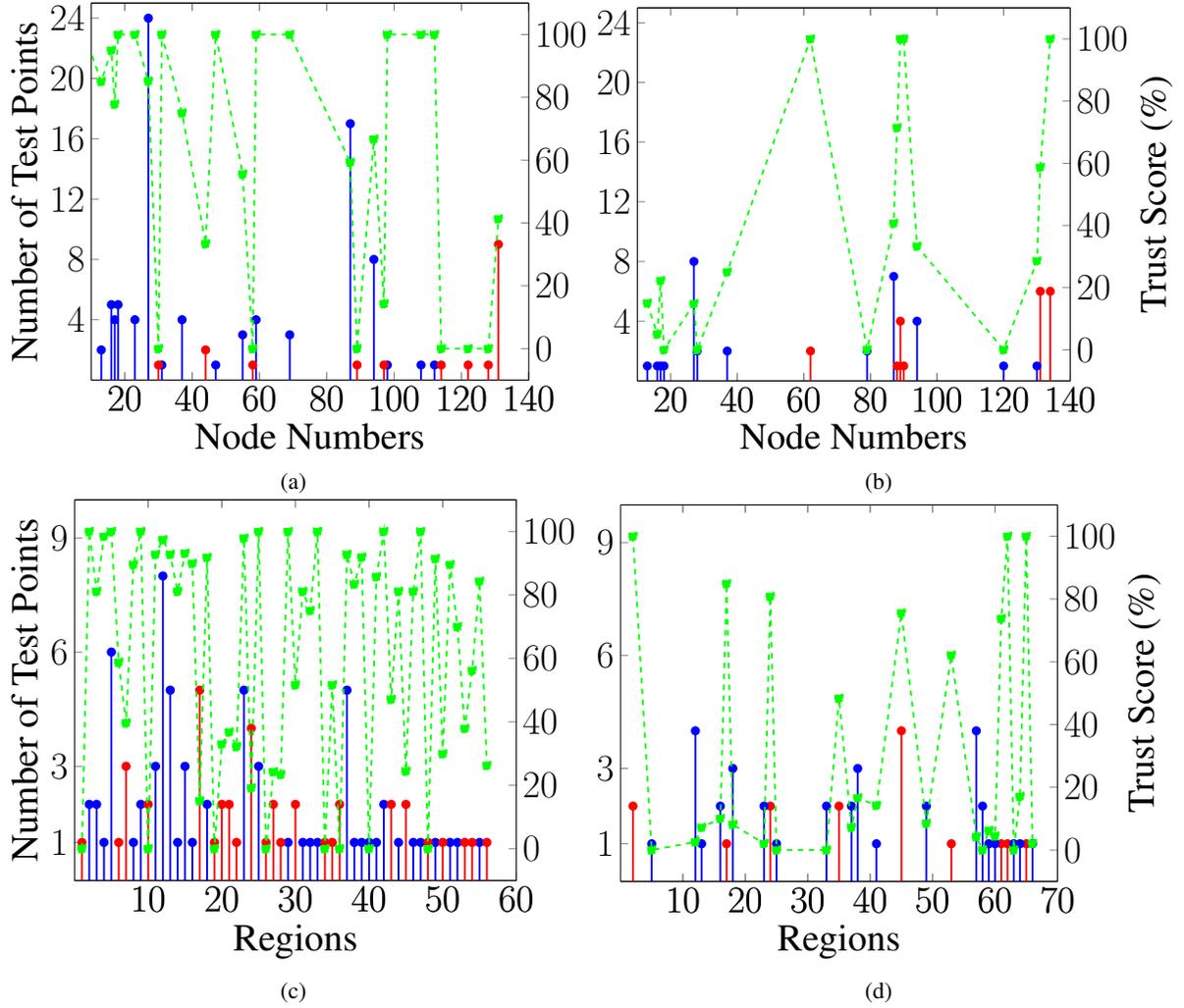
\begin{figure}[thpb!]
		\centering
        
        \begin{subfigure}[b]{0.48\columnwidth}
		  \centering
        \resizebox{1.02\columnwidth}{!}{
        \begin{tikzpicture}
		  \pgfplotsset{set layers,every tick label/.append style={font=\huge},label style={font=\huge}}
			
			\begin{axis}[
				scale only axis,
				xmin = 10, xmax = 140,
                ymin = 0, ymax = 25,
				xlabel = Node Numbers, 
				ylabel = Number of Test Points,
				ytick = {4,8,12,16,20,24,28},
                xtick = {1,20,40,60,80,100,120,140},
				axis y line*=left,
				legend style = {draw=none}]
				\addplot+ [blue, ycomb, mark=*, mark size = 2 pt, mark options={fill=blue}, line width = 1 pt] table {Data/TestSetA_DT_correct.dat};
                \addplot+ [red, ycomb, mark=*, mark size = 2 pt, mark options={fill=red}, line width = 1 pt] table {Data/TestSetA_DT_wrong.dat};
			\end{axis}
        
			\begin{axis}[
				scale only axis,
				xmin = 10, xmax = 140,
				xtick = {1,20,40,60,80,100,120,140},
				axis y line*=right,
				legend style = {draw=none}]
				\addlegendimage{/pgfplots/refstyle=plot_NumPoints}
				\addplot+ [green, dashed, sharp plot, mark=square*, mark size = 2 pt, mark options={fill=green}, line width = 1 pt] table {Data/TestSetA_DT_prob.dat};
			\end{axis}
		\end{tikzpicture}
        }
        \caption{\label{plot_NumPointsA_DT}}
        \end{subfigure}%
        \begin{subfigure}[b]{0.48\columnwidth}
        \centering
        \resizebox{1.02\columnwidth}{!}{
        \begin{tikzpicture}
		  \pgfplotsset{set layers,every tick label/.append style={font=\huge},label style={font=\huge}}
			
			\begin{axis}[
				scale only axis,
				xmin = 10, xmax = 140,
                ymin = 0, ymax = 25,
				xlabel = Node Numbers, 
				ytick = {4,8,12,16,20,24,28},
                xtick = {1,20,40,60,80,100,120,140},
				axis y line*=left,
				legend style = {draw=none}]
				\addplot+ [red, ycomb, mark=*, mark size = 2 pt, mark options={fill=red}, line width = 1 pt] table {Data/TestSetB_DT_correct.dat};
                \addplot+ [blue, ycomb, mark=*, mark size = 2 pt, mark options={fill=blue}, line width = 1 pt] table {Data/TestSetB_DT_wrong.dat};
								
			\end{axis}
   
			\begin{axis}[
				scale only axis,
				ylabel = Trust Score (\%),
				xmin = 10, xmax = 140,
				xtick = {1,20,40,60,80,100,120,140},
				axis y line*=right,
				legend style = {draw=none}, legend pos=north west]
				\addlegendimage{/pgfplots/refstyle=plot_NumPoints}
				\addplot+ [green, dashed, sharp plot, mark=square*, mark size = 2 pt, mark options={fill=green}, line width = 1 pt] table {Data/TestSetB_DT_prob.dat};
			\end{axis}
		\end{tikzpicture}
        }
        \caption{\label{plot_NumPointsB_DT}}
        \end{subfigure} \\
        \begin{subfigure}[b]{0.48\columnwidth}
        \centering
		  \resizebox{1.0\columnwidth}{!}{
        \begin{tikzpicture}
		  \pgfplotsset{set layers,every tick label/.append style={font=\huge},label style={font=\huge}}
			
			\begin{axis}[
				scale only axis,
				xmin = 0, xmax = 60,
                ymin = 0, ymax = 10,
				xlabel = Regions, 
				ylabel = Number of Test Points,
				ytick = {1,3,6,9},
				xtick = {10,20,30,40,50,60},
				axis y line*=left,
				legend style = {draw=none}]
				\addplot+ [blue, ycomb, mark=*, mark size = 2 pt, mark options={fill=blue}, line width = 1 pt] table {Data/TestSetA_adult_correct.dat};
                \addplot+ [red, ycomb, mark=*, mark size = 2 pt, mark options={fill=red}, line width = 1 pt] table {Data/TestSetA_adult_wrong.dat};
			\end{axis}
   
			\begin{axis}[
				scale only axis,
				xmin = 0, xmax = 60,
				xtick = {10,20,30,40,50,60},
				axis y line*=right,
				legend style = {draw=none}]
				\addlegendimage{/pgfplots/refstyle=plot_NumPoints}
				\addplot+ [green, dashed, sharp plot, mark=square*, mark size = 2 pt, mark options={fill=green}, line width = 1 pt] table {Data/TestSetA_adult_trust.dat};
			\end{axis}
        \end{tikzpicture}
        }
        \caption{\label{plot_NumPointsASEPC}}
        \end{subfigure}%
        \begin{subfigure}[b]{0.48\columnwidth}
        \resizebox{1.0\columnwidth}{!}{
		  \begin{tikzpicture}
		  \pgfplotsset{set layers,every tick label/.append style={font=\huge},label style={font=\huge}}
			
			\begin{axis}[
				scale only axis,
				xmin = 0, xmax = 70,
                ymin = 0, ymax = 10,
				xlabel = Regions, 
				ytick = {1,3,6,9},
				xtick = {10,20,30,40,50,60,70,80},
				axis y line*=left,
				legend style = {draw=none}]
				\addplot+ [red, ycomb, mark=*, mark size = 2 pt, mark options={fill=red}, line width = 1 pt] table {Data/TestSetB_adult_correct.dat};
                \addplot+ [blue, ycomb, mark=*, mark size = 2 pt, mark options={fill=blue}, line width = 1 pt] table {Data/TestSetB_adult_wrong.dat};		
			\end{axis}
    
			\begin{axis}[
				scale only axis,
				ylabel = Trust Score (\%),
				xmin = 0, xmax = 70,
				xtick = {10,20,30,40,50,60,70,80},
				axis y line*=right,
				legend style = {draw=none}] 
				\addlegendimage{/pgfplots/refstyle=plot_NumPoints}
				\addplot+ [green, dashed, sharp plot, mark=square*, mark size = 2 pt, mark options={fill=green}, line width = 1 pt] table {Data/TestSetB_adult_trust.dat};
			\end{axis}
        \end{tikzpicture}
        }
        \caption{\label{plot_NumPointsBSEPC}}
        \end{subfigure}
        
		\caption{Results of classifying the Adult Dataset: (\ref{plot_NumPointsA_DT}): DT for label $L_{+1}$; (\ref{plot_NumPointsB_DT}): DT for label $L_{-1}$; (\ref{plot_NumPointsASEPC}): SEP-C for label $L_{+1}$; (\ref{plot_NumPointsBSEPC}): SEP-C for label $L_{-1}$. In Figs.~\ref{plot_NumPointsA_DT} and \ref{plot_NumPointsASEPC}, the blue bars indicate the number of points in the respective regions that match the corresponding ground truth label $L_{+1}$ and those marked in red are misclassifications, while in Figs.~\ref{plot_NumPointsB_DT} and \ref{plot_NumPointsBSEPC}, red bars match the ground truth label $L_{-1}$ and blue bars indicate misclassifications. The green lines indicate the trust in classification.} 
        \label{fig:DTSEPC_AdultData}
\end{figure}

The results of the performance of both classifiers are shown in Fig.~\ref{fig:DTSEPC_AdultData}. Consider Figs.~\ref{plot_NumPointsASEPC} and \ref{plot_NumPointsBSEPC}. The $X-$axis in these figures indicate the unique regions in the feature space in which the test points lay. These regions could be single MVEs or an intersection of several MVEs of both labels as well. For example, region 61 is formed by an intersection of MVEs of both labels and which contains 1 record with label $L_{+1}$ and 5 records with label $L_{-1}$. Note that there may be many more such regions formed by the MVEs and their intersections. The points in the test set lie in 66 unique regions; the left $Y-$axis indicates the number of test points in that region.  The dotted lines in these figures show the trust score (marked by the $Y-$axis on the right) that is computed for each region according to \eqref{Eq:PostProbLabel}. Thus, a test point in region 61 is assigned a trust score of 73.7\% for label $L_{-1}$; the ground truth for this test point is actually label $L_{+1}$. Similarly, in Figs.~\ref{plot_NumPointsA_DT} and \ref{plot_NumPointsB_DT}, the $X-$axis lists the leaves in which the test points are found and the $Y-$axis on the right shows the ratio of points of one label to the total number of points from the training set found in that leaf; a trust score of 100\% shows that that leaf contains points of the same label. 

From an accuracy point of view - ratio of number of correct classifications to the number of testing records - DTC yielded an accuracy of 69.5\%, while SEP-C produced 52.5\%. The SVM classifier with linear and RBF kernel provided 71\% and 68\% accuracy, respectively. While these numbers may indicate that SEP-C performs poorly than these classifiers in terms of accuracy, the role of the Bayesian trust score in accepting the predicted labels should be highlighted. 

Consider Fig.~\ref{plot_NumPointsASEPC}, which shows the prediction made by SEP-C for test points with ground truth belonging to the majority class, that is, label $L_{+1}$. Nearly all points whose labels have been predicted correctly also have a high trust score, while those that are misclassified have a low trust score (less than 50\%). A similar result can be observed in Fig.~\ref{plot_NumPointsBSEPC} as well - most of the misclassified points (minority ground truth label) have a low trust score and the few that are correctly classified have a trust of 100\%. These imply that SEP-C is able to find regions in the feature space, however few, that contain points of the same label. The variations in the trust score for the different regions portray the overlapping landscape of the training data - a clean dataset would have few regions with nearly constant trust scores for all regions. If the trust score is viewed as a non-conformity measure, say with threshold $\epsilon=0.05$, then the labels of only those points whose trust score is greater than 95\% need be accepted. In this case, nearly all those points that are misclassified will be rejected; thus, if accuracy is measured for only the accepted test points, then naturally this score will be high for SEP-C. 

SEP-C also leads to misclassifications with high trust of some test points, for example, the red bars in regions 7 and 17 in Fig.~\ref{plot_NumPointsASEPC}. None of these points actually lay in any of the MVEs or their intersection, thus satisfying \textbf{Case}~3, as described in Sec.~\ref{Sec:RulesClassify}. Now, these points are assigned a label by expanding the MVEs closest to them. For these test points, as these MVEs belonged to the points with the minority label, they were also assigned the minority label. These results highlight the ``strangeness'' of these test points; had they been close to other points of the same label - making them ``less strange '' - they would have been contained inside one of the MVEs anyway. Thus, the prediction of the label of a test point made by SEP-C can also be rejected if the point in question is not contained within one of the MVEs found by SEP-C. Further, if the number of such misclassifications exceeds $n_{\text{imp}}$ in one region, then, these test points, with their correct labels, can be included in the training set and SEP-C can be used to partition the dataset afresh. 

\subsection{WDBC Dataset}

This dataset consists of 569 records with 357 label $L_{+1}$ (\textit{Benign}) records and 212 records with label $L_{-1}$ (\textit{Malignant}); each record is defined by 30 features. One of the earliest works on this dataset, \cite{street1993nuclear}, reports 97\% accuracy with the application of a classifier based on the Multi-Surface Method (MSM), known as the MSM-Tree (MSM-T) classifier; this accuracy was obtained following ten-fold cross-validation on three of the thirty features, namely, \texttt{mean texture}, \texttt{worst area}, and \texttt{worst smoothness}. MSM-T solves a linear program to split at a node and recursively identifies several regions, based on separating hyperplanes, in the feature space; a DTC-type method is then used to predict a label of a test point. In \cite{saygili2018classification}, the authors build classification models using SVM, k-NN, Naive Bayes, J48, Random Forest, and Multilayer Perceptron methods. The dataset was preprocessed and 10-fold cross validation was applied. The Random Forest classifier is shown as the the most accurate model, with an accuracy of 98.77\%. The Linear SVM classifier with preprocessing gave 98.07\% accuracy.

In \cite{salama2012breast}, a comparison of classification accuracy among Decision Tree, Multi-Layer Perceptron, Naive-Bayes, Sequential Minimal Optimization (SMO), and Instance-based k-NN methods is presented for WDBC dataset as well as original and Prognostic datasets. The aim is to find the best combination of the algorithms in order to improve accuracy of the model. The proposed SMO uses the PCA and $\chi$-squared test for feature selection and it is observed that this had little effect on performance of the combination of algorithms on the WDBC dataset. The highest accuracy achieved by the SMO is 97.7\%, while other ensembles have report lower accuracy. In \cite{aalaei2016feature}, the use of an ANN, Particle Swarm Optimization and Genetic Algorithm shows a slight improvement in accuracy when 14 out of 32 features were selected during feature selection, with accuracy of 97.3\%, 97.2\% and 96.6\% respectively.

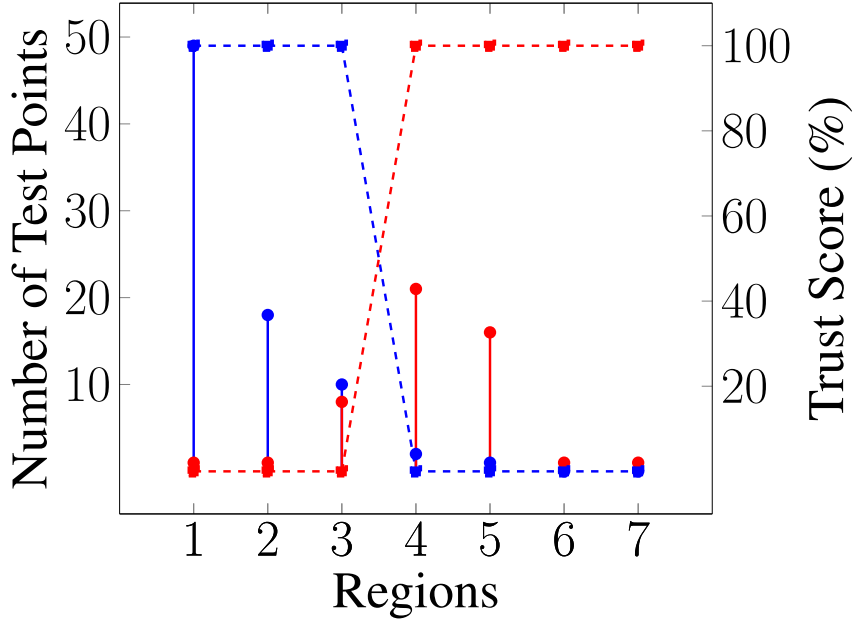
\begin{figure}[thpb!]
		\centering
		\resizebox{0.7\columnwidth}{!}{
		
		\begin{tikzpicture}
		  \pgfplotsset{set layers,every tick label/.append style={font=\huge},label style={font=\huge}}
			
			\begin{axis}[
				scale only axis,
				xmin = 0, xmax = 8,
				xlabel = Regions, 
				ylabel = Number of Test Points,
				ytick = {10,20,30,40,50},
				xtick = {1,2,3,4,5,6,7},
				axis y line*=left,
                ]
				\addplot+ [blue, ycomb, mark=*, mark size = 2 pt, mark options={fill=blue}, line width = 1 pt] table {Data/TestSetAF4_WDBC_correct.dat};
                \addplot+ [blue, ycomb, mark=*, mark size = 2 pt, mark options={fill=blue}, line width = 1 pt] table {Data/TestSetAF4_WDBC_wrong.dat};
                \addplot+ [red, ycomb, mark=*, mark size = 2 pt, mark options={fill=red}, line width = 1 pt] table {Data/TestSetBF4_WDBC_correct.dat};
                \addplot+ [red, ycomb, mark=*, mark size = 2 pt, mark options={fill=red}, line width = 1 pt] table {Data/TestSetBF4_WDBC_wrong.dat};	
			\end{axis}
   
			\begin{axis}[
				scale only axis,
				ylabel = Trust Score (\%),
				xmin = 0, xmax = 8,
				xtick = {1,2,3,4,5,6, 7},
				ytick = {20,40,60,80,100},
				axis y line*=right,
				legend style={at={(1,0.7)}, {draw=none},{font=\Large}}]
				
				\addplot+ [blue, dashed, sharp plot, mark=square*, mark size = 2 pt, mark options={fill=blue}, line width = 1 pt] table {Data/TestSetA_trust_WDBC.dat};
                \addplot+ [red, dashed, sharp plot, mark=square*, mark size = 2 pt, mark options={fill=red}, line width = 1 pt] table {Data/TestSetB_trust_WDBC.dat};
								
			\end{axis}
		\end{tikzpicture}
		}
		\caption{SEP-C Trust scores - WDBC dataset} 
        \label{fig:SEPC_WDBC}
	\end{figure}
   
The results of applying SEP-C on this dataset are shown in Fig.~\ref{fig:SEPC_WDBC}; prior to its implementation, it is found that the points of the two labels have almost no overlap, thus indicating that a high accuracy in classification is possible. A 4-fold, 90-10 train-test split-cross validation technique was used with all 30 features and $n_{\text{Imp}}$ = 2. In each fold, SEP-C finds 3 partitions for each class; these are Regions 1-3 for training set points with label $L_{+1}$ and Regions 4-7 for points with label $L_{-1}$. Region 7 is formed by the intersection of 1\textsuperscript{st} and 2\textsuperscript{nd} ellipsoids of label $L_{-1}$.  As can be seen in Fig.~\ref{fig:SEPC_WDBC}, except for about 10 test points with ground truth label $L_{-1}$, that lie in Region 3 and marked in red, all other points are correctly classified. The same holds true for test points lying in Region 4-7 - the predictions with trust score of 100\% matches the ground truth label. The points in red and lying in Region 3, which are misclassified with high trust, deserve a closer look. During the validation stage of SEP-C, the labels of each of these test points is predicted one-by-one and then removed from the train-test set. Thus, since Region 3 contains points with label $L_{+1}$, any test point that is contained in this region will also get this label, according to Rule~1 of SEP-C. Now, when all of these test points are considered together, the results show that they form a cluster that can be interpreted as a small disjunct. If the number of permitted misclassfications is chosen as less than 10, the number shown in Fig.~\ref{fig:SEPC_WDBC}, partitioning will need to be performed again. Now, SEP-C will find a new partition for these points. 

\subsection{VC Dataset}

This dataset consists of 100 records with label $L_{+1}$ (\textit{Normal}) and 210 records with label $L_{-1}$ (\textit{Abnormal}). The MVEs of the points of individual labels from the entire dataset show OVR values, with the MVE formed in the region of overlap, of 0.67 for label $L_{+1}$ and 0.006 for label $L_{-1}$. In fact, it is only 7 out of 100 records of points with label $L_{+1}$ that lie outside the region of overlap while 157 points with label $L_{-1}$ lie outside the region of overlap. Such an unequal distribution of points indicate a possibility of misclassification of points with ground truth $L_{+1}$ - this observation is also supported when SEP-C is applied on this dataset. By performing a 10-fold cross validation with $n_{\text{Imp}} = 2$ led to accuracy values ranging from as low as 64.2\% for one fold to as high as 92.8\% for another. In \cite{Handayani_2019}, a prediction accuracy of 83\% with k-NN algorithm is reported for this dataset. A comparative study in \cite{Rizki2015} gives accuracy as 89.03\% for k-NN, 88.06\% for ANN, and 86.13\% for Naïve Bayes with cross validation, after applying genetic algorithms and bagging technique. 

When the variation in the accuracy across folds is analysed for SEP-C, it is seen that in the worst performing case, several points with ground truth label $L_{+1}$ - also the minority class in the region of overlap - lie in MVEs that contain points only with label $L_{-1}$. On the other hand, the same points form their own MVE in the fold when the accuracy is the highest. Note that with an increase in misclassification in the worst performing fold indicates a need to repartition the dataset with the correct labels. Thus, with this repartitioning, it is likely that these points would have formed their own MVE. A similar result is seen in the WDBC dataset for points that lie in Region 3 that are misclassified, as shown in Fig.~\ref{fig:SEPC_WDBC}.

\subsection{Multiclass Classification - Iris Dataset}

\begin{figure}[htpb!]
    \centering
    \includegraphics[width=0.8\columnwidth]{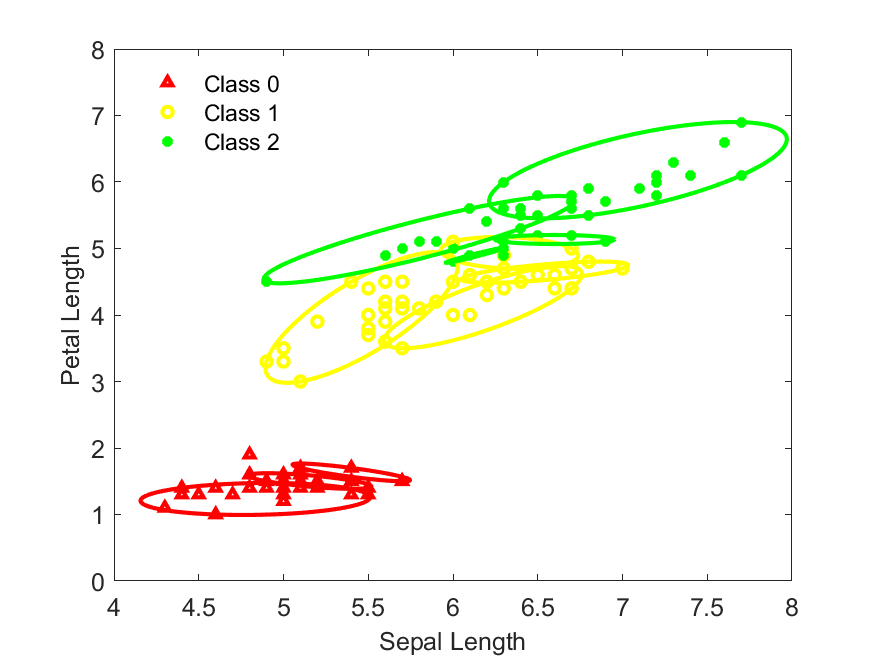}
    \caption{SEP-C Classification for Iris Dataset}
    \label{Fig:Iris_2feat}
\end{figure}

SEP-C, when applied to the multi-class Iris dataset, results in the MVEs shown in Fig.~\ref{Fig:Iris_2feat}; the one-vs-all approach is followed and 2 features - \texttt{sepal length} and \texttt{petal length} are selected. The class labels for the 3 classes are assigned as - class 0 (\textit{setosa}), class 1 (\textit{versicolor}), and class 2 (\textit{verginica}). With 2 and all 4 features and $n_{\text{Imp}}=2$, 3 MVEs were obtained for class 0 and 4 MVEs each for classes 1 and 2. The test points of class 0 were classified accurately. As can be seen, there is significant overlap between the points of classes 1 and 2, hence, SEP-C also misclassified points of class 1. 

The performance of SEP-C on the Iris dataset is presented as a comparison with the GMMs presented in \cite[Sec.~8.4]{McLachlan2019} for this dataset. The MVEs obtained from SEP-C, which are in essence Gaussians, are a natural result of the iterative RCH function that is an integral part of SEP-C. Thus, there is no need to specify the number or type of Gaussian mixtures in SEP-C to cluster the 3 classes.

\section{Conclusions}

This paper proposes a novel, convex optimization-based, supervised classifier, SEP-C, for binary and multi-class classification. The algorithm does not require prior knowledge about the nature of the dataset and its underlying distribution and in addition, its iterative characteristic lends it the property of being free of hyperparameters, such as used in SVMs. The single user-defined parameter is the number of misclassifications permitted in the partitioning, which we believe has a greater intuitive appeal than hyperparameters defined in the cost functions. Rules of classification can be stated based on the obtained partitions and based on the outputs of these rules, a Bayesian trust score in the prediction can also be calculated. This trust score, which allows for a user to accept or reject the predicted label, lends a degree of explainability to the prediction, as it is based on the strangeness of the test point to other points in a local region of the dataset. While the accuracy comparisons made in this paper are for the entire test set, when the predicted and ground truth labels are compared for individual points and it emerges that a set of points have been misclassified consistently and whose number is greater than the user-defined allowed value, then, this is an indicator that SEP-C should be applied on the updated dataset for fresh partitioning. As we have shown in the results, small disjuncts - which exhibit such properties - can be identified in the retraining of SEP-C. 

The presented results also highlight that SEP-C can act on datasets of different types - containing only categorical or continuous features or their mix. The multiple hyperplanes that are determined in each iteration of the RCH function make it applicable for datasets that are not linearly separable. Thus, there is no need to search for a suitable kernel or an appropriate transform that would render the data to become linearly separable. SEP-C can be extended to identification of small disjuncts; for datasets subjected to a data shift; as well as for time-series data. Further, the density of points in the determined MVEs can be analysed further to examine the need for cross-fold validations and selecting the fold that best partitions the dataset. 

\section*{Acknowledgments}
Ranjani Niranjan would like to thank Prateeksha Foundation for providing financial support for her doctoral program at IIIT-Bangalore.

\bibliographystyle{unsrt}  
\bibliography{references}

\end{document}